\documentclass[runningheads]{llncs}

\usepackage{eccv}

\usepackage{eccvabbrv}
\usepackage{graphicx}
\usepackage{booktabs}

\usepackage[accsupp]{axessibility}  
\graphicspath{{figures/}}
\usepackage[table,xcdraw,dvipsnames]{xcolor} 

\usepackage{tcolorbox}

\usepackage{hyperref}

\usepackage{orcidlink}

\begin{document}

\title{ZeroSplat: Generalized Referring Segmentation in 3D Gaussian Splatting} 

\titlerunning{ZeroSplat}

\author{
	Jiayu Ding\inst{1,4}\thanks{Equal contribution. \textsuperscript{\dag} Corresponding author.} \and
	Meilu Song\inst{2}\textsuperscript{*} \and
	Xiaoyi Zhang\inst{3}
	\\
	Hongbo Jin\inst{1,4} \and
	Yichen Jin\inst{1} \and
	Xiangtian Si\inst{3}\textsuperscript{\dag}
}

\authorrunning{J.~Ding et al.}

\institute{
	\textsuperscript{1}Peking University \quad
	\textsuperscript{2}North China Electric Power University \\
	\textsuperscript{3}China University of Geosciences \quad
	\textsuperscript{4}InkMind.AI
}
\maketitle

\begin{abstract}
	Recent advancements in 3D Gaussian Splatting (3DGS) have enabled language-guided scene understanding. However, existing Referring 3D Gaussian Splatting  (R3DGS) methods are fundamentally restricted to single-target queries. To reflect the ambiguity of real-world instructions, we introduce the Generalized Referring 3D Gaussian Splatting Segmentation (GR3DGS) task, which requires dynamically segmenting an arbitrary number of targets (0, 1, or $N$). To facilitate comprehensive evaluation of this new task, we construct two new benchmarks: GR-LERF and GR-ScanNet. Crucially, existing R3DGS paradigms exhibit fundamental technical bottlenecks that severely limit their performance on the GR3DGS task: they lack intrinsic 3D point-level understanding by operating merely on 2D rendered pixels, and they incur prohibitive computational overhead by requiring per-scene optimization to embed heavy semantic features. To dismantle these bottlenecks, we propose ZeroSplat, a novel training-free and zero-feature framework. ZeroSplat lifts 2D Vision-Language Model (VLM) priors into 3D space through robust multi-view geometric constraints. This strategy enables intrinsic point-level understanding without incurring any additional feature storage. Extensive experiments demonstrate that ZeroSplat significantly outperforms state-of-the-art methods across generalized and single-target scenarios while maintaining exceptional efficiency. \textit{Project Page: https://inkmind-ai.github.io/ZeroSplat}
\end{abstract}

\section{Introduction}

3D Gaussian Splatting (3DGS)~\cite{kerbl_3d_2023} represents scenes using explicit 3D Gaussians to enable high-quality real-time rendering. Beyond visual synthesis, recent research has evolved to equip 3DGS with semantic understanding capabilities. Initially, Open-Vocabulary 3DGS Understanding methods~\cite{li_instancegaussian_2025,wu2024opengaussian,liang2024supergseg,sun2025cags,yin2025semantic,jun-seong_dr_2025,jiang2025votesplat,marrie2025LUDVIG} distilled 2D foundation model features into 3D space, allowing users to query scenes using text. However, these approaches typically rely on fixed category names or simple noun phrases. 
This limitation impedes free-form language understanding, core to Embodied AI~\cite{li2026lmm,kim2024openvla,ding20263did,tang2026video,li2026egocentric} and multimodal LLM agent~\cite{xiao2025visual,xiao2026promptbased,jin2026himachierarchicalmacromicrolearning,jin2026context,chen2026physics}, with user queries carrying fine-grained attributes and intricate spatial relations.
Hence, Referring 3D Gaussian Splatting (R3DGS)~\cite{ReferSplat} fills this gap by segmenting targets from elaborate text prompts.

\begin{figure*}[t!]
	\centering
	\includegraphics[width=1\linewidth]{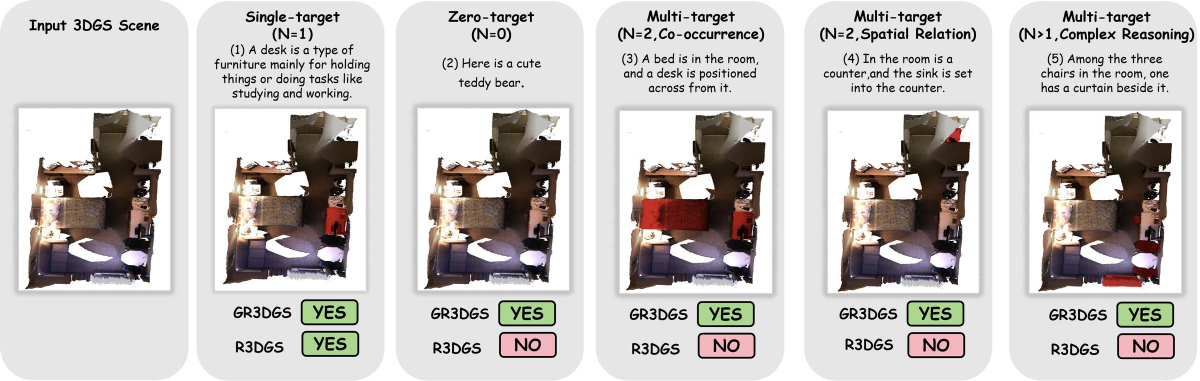}
	\caption{Traditional R3DGS is limited to single-target cases(1). In contrast, GR3DGS can handle scenarios with any number of targets, including no target (2), single target, and multiple targets (3-5). }
	\label{001}
\end{figure*}

However, existing R3DGS paradigms~\cite{ReferSplat} suffer from a critical limitation: they are strictly confined to a ''single-target'' setting. As illustrated in Fig.~\ref{001}, these methods presuppose that each instruction corresponds to exactly one target in the scene. 
This assumption severely hinders flexibility in real-world scenarios, where user instructions are inherently uncertain, often involving multi-target requests (e.g., find all red chairs) or no-target queries (i.e., the object is absent).
To address this, we introduce the \textbf{Generalized Referring 3D Gaussian Splatting Segmentation (GR3DGS)} task. This task mandates parsing instructions to segment an arbitrary number of targets (0, 1, or $N$), imposing higher demands on semantic discrimination and robustness against ambiguity. To facilitate evaluation, we construct \textbf{GR-LERF} for pixel-level assessment and \textbf{GR-ScanNet} for intrinsic point-level assessment via 3D annotations.

Beyond the task formulation, the existing technical paradigm encounters significant bottlenecks when applied to GR3DGS, suffering from two critical limitations:
\textbf{(i) Lack of 3D Point-Level Understanding:} The current R3DGS method utilizes 2D rendered pixels as the fundamental unit for semantic processing rather than operating on discrete 3D Gaussian points. Fundamentally, this approach remains confined to 2D image-level understanding, failing to leverage the explicit 3D geometric structure of the scene to resolve complex spatial ambiguities.
\textbf{(ii) Prohibitive Computational and Memory Overheads:} Furthermore, the existing method necessitates prolonged scene-specific optimization and requires embedding high-capacity semantic features into millions of Gaussian points. This massive storage footprint and peak memory consumption render it highly impractical for real-time applications and scalable deployment on resource-constrained devices.

To bridge these gaps, we propose \textbf{ZeroSplat}, a novel training-free and zero-feature framework tailored for GR3DGS. Diverging from the prevailing paradigm of coupling additional semantic parameters, ZeroSplat is built on the core insight that robust 3D semantic understanding does not mandate altering the intrinsic scene representation. Instead, it can be achieved by lifting 2D foundation model priors into 3D space through geometric constraints. 
By projecting semantic cues from 2D Vision-Language Models (VLMs) onto the 3D structure, our method directly filters and localizes targets within the original set of 3D Gaussians. 

In summary, our contributions are as follows:
\begin{itemize}
	\item We introduce a new task termed Generalized Referring 3D Gaussian Splatting  Segmentation (GR3DGS).
	\item To support future research in GR3DGS, we construct GR-LERF and GR-ScanNet for evaluation at both pixel and point levels.
	\item To address GR3DGS challenges, we propose ZeroSplat, a zero-feature and training-free framework achieving intrinsic point-level understanding.
	\item Experiments show that our method outperforms existing approaches in both generalized and single-target scenarios.
\end{itemize}

\section{Related Works}
\noindent\textbf{Preliminary: 3D Gaussian Splatting}
3D Gaussian Splatting (3DGS)~\cite{kerbl_3d_2023} represents 3D scenes using a set of explicit 3D Gaussians $\mathcal{G} = \{g_i\}_{i=1}^N$. Each Gaussian $g_i$ is characterized by its mean position, covariance matrix (controlling scale and orientation), color, and opacity. To render a 2D image, these Gaussians are projected onto the image plane and blended in a depth-sorted order. The final color $C(p)$ for a pixel $p$ is computed through alpha compositing~\cite{munkberg2022extracting}:
\begin{equation}
	\label{eq:rendering}
	C(p) = \sum_{i=1}^{|\mathcal{G}_p|} c_i \alpha_i \prod_{j=1}^{i-1} (1 - \alpha_j),
\end{equation}
where $c_i$ and $\alpha_i$ denote the color and effective opacity of the $i$-th Gaussian intersecting pixel $p$. The effective opacity $\alpha_i$ is the product of the learned opacity and the spatial influence of the projected Gaussian. The term $T_i = \prod_{j=1}^{i-1} (1 - \alpha_j)$ represents the accumulated transmittance, accounting for light attenuation from all preceding Gaussians along the viewing ray.

\noindent\textbf{Language-grounded 3D Gaussian Splatting}
Following the success of 3D Gaussian Splatting (3DGS), recent studies have integrated open-vocabulary and natural language understanding into the 3DGS framework. Existing approaches primarily follow two paradigms: pixel-based and point-based methods.
Pixel-based methods adopt a ``render-then-match'' strategy, where the scene is first rendered into dense 2D feature maps for semantic reasoning in the image space. Several works~\cite{shi_language_2024,zhou_feature_2024-1,qin_langsplat_2024,ye_gaussian_2024,qu_goi_2024,ReferSplat} investigate open-vocabulary 3DGS understanding by distilling semantic features from 2D foundation models into view-consistent 3D representations to enable efficient semantic rendering. More recently, some studies have explored referring 3DGS understanding. For example, ReferSplat~\cite{ReferSplat} aligns 3D Gaussians with text queries using a position-aware cross-modal module to improve spatial reasoning. Although it builds an explicit 3D referring field, its localization still relies on identifying target pixels within rendered 2D images. Consequently, these methods treat 3D representations primarily as a proxy for rendering. They lack structural understanding of the scene and cannot directly identify or manipulate individual Gaussian primitives, limiting their application in interactive tasks~\cite{liu2026egotl,jin2026vistamitigatingsemanticinertia,jin2026contextguard}.
To address these limitations, point-based methods employ a ``match-then-render'' paradigm, treating 3D Gaussian primitives as the basic units of understanding. For open-vocabulary tasks, methods such as OpenGaussian~\cite{wu2024opengaussian} and InstanceGaussian~\cite{li_instancegaussian_2025}  use 2D masks from models like SAM to learn 3D-consistent instance features. Other works, including Dr.Splat~\cite{jun-seong_dr_2025}, LUDVIG~\cite{marrie2025LUDVIG} and ExtrinSplat~\cite{ding2026extrinsplat}, lift 2D features directly onto 3D Gaussian points via feature aggregation or graph diffusion. While point-based methods have made progress in open-vocabulary scenarios, extending their advantages to referring 3DGS understanding remains largely unexplored.

\noindent\textbf{Referring Segmentation}
Referring segmentation aims to localize target regions described by natural language queries.
Initially developed in the 2D domain, Referring Image Segmentation~\cite{liu2017recurrent,wang2022cris} (RIS) has achieved remarkable success by aligning linguistic features with pixels.
However, RIS remains confined to the image plane, lacking the spatial reasoning capabilities essential for real-world interaction.
To address this limitation, 3D Referring Segmentation~\cite{huang2021text,he2024refmask3d,wu2024rg,wu20243d,liu2024less} (3D RES) extends language grounding to 3D data.
Recently, to better reflect real-world complexities, the task has evolved into Generalized 3D Referring Segmentation~\cite{wu20243d} (3D-GRES).
Breaking the constraint of single-object localization, 3D-GRES allows expressions to refer to an arbitrary number of targets, aiming to predict a binary mask covering all relevant 3D points.
Despite these advancements, existing methods and datasets rely primarily on 2D images or 3D point clouds. Consequently, they cannot be applied to GR3DGS.

\section{Method}

\subsection{Task Definition and Method Overview}
Formally, given a 3D scene reconstructed from a set of multi-view images $\mathcal{I} = \{I_i\}_{i=1}^N$ and represented by a 3DGS field $\mathcal{G}$, along with a free-form natural language expression $\mathcal{T}$, the GR3DGS task aims to assign a binary semantic label to each Gaussian $g \in \mathcal{G}$. Unlike standard R3DGS, which strictly assumes that $\mathcal{T}$ corresponds to exactly one target, GR3DGS formulates a more realistic open-world setting. The expression $\mathcal{T}$ can refer to an arbitrary number of instances. Consequently, the model must dynamically segment multiple targets, a single target, or output an empty mask if the queried object is absent from the scene. This setting requires the framework to perform precise 3D spatial reasoning while maintaining robust semantic discrimination to reject false positives.

\begin{figure*}[t!]
	\centering
	\includegraphics[width=1\linewidth]{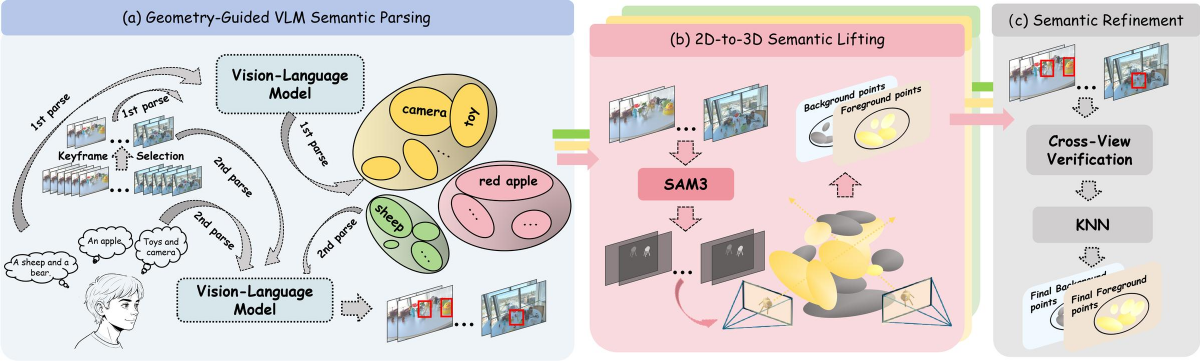}
	\caption{Overview of our method. (a) Keyframes are first extracted from the input scene. A VLM then performs a two-stage parsing: the first extracts semantic labels from text via multi-view keyframe analysis, while the second localizes targets with 2D bounding boxes using these labels, referring text, and keyframes. (b) Guided by these labels, partial multi-view 2D masks are extracted across all scene views and used to back-project objects into 3D point groups. (c) Finally, erroneous Gaussians are filtered using the VLM-predicted 2D boxes, and internal 3D Gaussian structures are filled via a KD-Tree-based KNN spatial label diffusion algorithm.}
	\label{002}
\end{figure*}

To tackle GR3DGS, we propose a three-stage end-to-end framework that hierarchically connects unstructured text to 3D geometric anchors (Figure~\ref{002}).
First, we select geometric keyframes via curvature evaluation and use multi-stage interactions with a Vision-Language Model (VLM) to extract semantic labels and 2D localizations from the text (Sec. 3.2). Second, we use SAM3 to generate 2D semantic masks and lift them into the 3D Gaussian field (Sec. 3.3). Finally, we apply multi-view verification and a spatial diffusion algorithm to refine the 3D semantic structure (Sec. 3.4).

\subsection{Geometry-Guided VLM Semantic Parsing}

The first step of our pipeline parses the free-form textual query into semantic and spatial priors. Directly segmenting scenes using complex instructions often introduces ambiguity. To address this, we extract a concise semantic label from the query to facilitate downstream 2D mask generation. However, this label denotes a broad category that includes both the intended target and irrelevant instances. To resolve this ambiguity, we employ a VLM to generate 2D bounding box anchors based on the original query. These anchors provide spatial constraints to accurately isolate the referred target from the broader semantic category.

\noindent \textbf{Geometry-Guided Keyframe Selection.} To apply the VLM efficiently, we avoid processing all $N$ frames of a video sequence, which causes high computational cost, latency, and view redundancy. Instead, we greedily select a compact subset $\mathcal{I}_{key}$ of $K$ keyframes based on scene geometric curvature. In 3D scenes, complex geometric regions such as edges and corners typically contain richer semantic information than flat areas. Given the explicit point cloud $\mathcal{P}$ formed by the Gaussian centers, we first perform sub-voxel downsampling on $\mathcal{P}$ to obtain a representative subset $\mathcal{P}_{sub}$ at a resolution of $\gamma s$, where $s$ denotes the base voxel size and $\gamma$ is the downsampling ratio. We then compute the eigenvalues $\lambda_1 \ge \lambda_2 \ge \lambda_3$ of the covariance matrix formed by neighbors within a search radius $r = \eta s$, where $\eta$ is a fixed coefficient. We define the local surface variation rate as $\sigma_{\mathbf{p}} = \lambda_3 / (\lambda_1 + \lambda_2 + \lambda_3 + \epsilon)$, where $\epsilon $ is a small constant to ensure numerical stability.  We then map $\sigma_{\mathbf{p}}$ to a saliency weight $w_{\mathbf{p}}$ using min-max normalization, scaling it to a target interval $[w_{min}, w_{max}]$:
\begin{equation}
	w_{\mathbf{p}} = w_{min} + (w_{max} - w_{min}) \frac{\sigma_{\mathbf{p}} - \sigma_{min}}{\sigma_{max} - \sigma_{min}}
\end{equation}

To ensure efficient scene coverage, we discretize the scene into a voxel grid. We define the weight $W(v)$ of each voxel $v$ as the maximum saliency weight $w_{\mathbf{p}}$ among its internal points. Let $\mathcal{U}$ be the set of voxels covered by the selected keyframes, initialized as $\mathcal{U} = \emptyset$. For each candidate frame $I_i$, we back-project its pixels into 3D space using the depth map and camera parameters to obtain the set of valid observed voxels $\mathcal{V}(I_i)$. We evaluate the value of adding a new frame using marginal gain:
\begin{equation}
	\text{Gain}(I_i | \mathcal{U}) = \sum_{v \in \mathcal{V}(I_i) \setminus \mathcal{U}} W(v)
\end{equation}

In each iteration, we add the frame with the maximum marginal gain to $\mathcal{I}_{key}$ and merge its visible voxels into $\mathcal{U}$, stopping when $K$ frames are selected. This strategy filters redundant background views, maximizing the coverage of high-value geometric regions while minimizing data redundancy.

\noindent \textbf{Hierarchical VLM Reasoning.} With the selected keyframes $\mathcal{I}_{key}$, we execute a two-stage VLM interaction to establish the aforementioned priors. The first stage performs semantic label extraction. We input the referring text $\mathcal{T}$ and the keyframes $\mathcal{I}_{key}$ into the VLM to extract a concise semantic label set $\mathcal{C} = \{C_1, C_2, \dots, C_M\}$ from the verbose description. The second stage performs 2D geometric localization. We feed the extracted label $C_j$, the original text $\mathcal{T}$, and the keyframes $\mathcal{I}_{key}$ back into the model. Prompted accordingly, the VLM outputs a normalized 2D bounding box for the target object in each keyframe. These bounding boxes act as crucial spatial priors for filtering 3D Gaussian artifacts in subsequent lifting stages.

\subsection{2D Semantic Mask Generation and 3D Lifting}

With the semantic and spatial priors established by the VLM, this module instantiates these 2D cues into a precise 3D segmentation. We first extract and adaptively fuse high-fidelity 2D masks using the extracted labels. Crucially, we then lift these 2D observations into the 3D Gaussian space by exploiting intrinsic volume rendering properties, and perform strict cross-view background cropping.

\noindent \textbf{Adaptive View Selection and Mask Fusion.} We employ an VLM to parse the referring expression into a concise semantic label set $\mathcal{C}=\{C_j\}_{j=1}^L$, containing $L$ semantic labels, which are then input into SAM3 to extract multi-view 2D masks. For each category $c_j$ and frame $I_i$, SAM3 generates multiple candidate masks. Given a text prompt, SAM3 generates multiple candidate masks for each frame $I_i$. Let $m_i^{(1)}$ and $m_i^{(2)}$ be the dominant mask with the highest confidence and the sub-optimal mask in frame $I_i$, with confidences $c_i^{(1)}$ and $c_i^{(2)}$ respectively. To prevent noisy masks in poor views from degrading 3D multi-view geometric consistency, we adaptively filter the image sequence. \noindent We define a high-confidence threshold $\tau_{high}$, a base threshold $\tau_{base}$, and a fallback threshold $\tau_{safe}$ to form candidate view sets $\mathcal{I}_{high} = \{I_i \mid c^{(1)}_i > \tau_{high}\}$, $\mathcal{I}_{base} = \{I_i \mid c^{(1)}_i > \tau_{base}\}$, and $\mathcal{I}_{safe} = \{I_i \mid c^{(1)}_i > \tau_{safe}\}$. To balance semantic purity and geometric coverage, we determine the final valid view set $\mathcal{I}_{sel}$ with quantity threshold $N_{target}$ and $N_{safe}$ as follows:
\begin{equation}
\mathcal{I}_{sel} = 
\begin{cases} 
\mathcal{I}_{high} & \text{if } |\mathcal{I}_{high}| \ge N_{\text{target}} \\
\text{Top}_{N_{\text{target}}}(\mathcal{I}_{base}) & \text{else if } |\mathcal{I}_{base}| > 0 \\
\text{Top}_{\min(|\mathcal{I}_{safe}|, N_{\text{safe}})}(\mathcal{I}_{safe}) & \text{otherwise}
\end{cases}
\end{equation}

This strategy ensures that we prioritize high-confidence views while maintaining a sufficient number of frames (up to $N_{target}$) for reconstruction. Next,  for each $I_i \in \mathcal{I}_{sel}$ and each category $C_j \in \mathcal{C}$, we generate a per-category mask $M_{i,j}$. Because SAM3 often over-segments objects, we merge the sub-optimal mask $m^{(2)}_i$ with the dominant mask $m^{(1)}_i$ if its confidence $c^{(2)}_i$ exceeds a strict threshold $\tau_{merge}$:
\begin{equation}
M_i = m_i^{(1)} \cup (\mathbb{I}(c_i^{(2)} > \tau_{merge}) \cdot m_i^{(2)})
\end{equation}

To aggregate the semantic label sets corresponding to each referring expression, the final semantic mask $\hat{M}_i$ for frame $I_i$ is obtained via a cross-category union:
\begin{equation}
\hat{M}_i = \bigcup_{j=1}^{L} M_{i,j}
\end{equation}

\noindent \textbf{Mask-Based Back-Projection Initialization.} To lift 2D semantics to 3D space, we leverage the volume rendering properties of 3DGS to assign semantic labels to individual Gaussians. 
First, we calculate the rendering contribution of a single Gaussian from a specific view. In standard forward rendering, the contribution weight $w(r, g_j)$ of the $j$-th Gaussian $g_j$ along ray $r$ is defined by its accumulated transmittance and opacity:
\begin{equation}
	w(r, g_j) = T(r, g_j) \alpha(r, g_j)
\end{equation}
where $T(r, g_j)$ is the accumulated transmittance before the ray reaches $g_j$, and $\alpha(r, g_j)$ is the opacity. We then aggregate the multi-view semantic responses for each Gaussian. For a Gaussian $g_j$, we check its projected pixel set $\mathcal{R}_i$ across all valid views $I_i \in \mathcal{I}_{sel}$. Using the 2D final mask value $M_i(r) \in \{0, 1\}$ at each pixel, we compute a global score $W_k(g_j)$ for $g_j$ being foreground ($k=1$) or background ($k=0$):
\begin{equation}
	W_k(g_j) = \sum_{I_i \in \mathcal{I}_{sel}} \sum_{r \in \mathcal{R}_i} \mathbb{I}(M_i(r) = k) w_i(r, g_j)
\end{equation}
Finally, we build the initial foreground 3D Gaussian set $\mathcal{G}_{fg}$ using hard assignment. A Gaussian is added to $\mathcal{G}_{fg}$ if its foreground score is strictly higher than its background score ($W_1(g_j) > W_0(g_j)$). Otherwise, it is classified as background.

\noindent \textbf{Cross-View Background Cropping. } To further improve the purity of the 3D semantic field, we design a cross-view background cropping mechanism. For each Gaussian $g$ currently in $\mathcal{G}_{fg}$, we compute its 2D projection across all valid views in $\mathcal{I}_{sel}$. We count the number of times it projects within the valid imaging area as $N_{val}(g)$, and the number of times it falls onto a background region as $N_{conflict}(g)$. We define the conflict ratio as:
\begin{equation}
    \rho_g = \frac{N_{conflict}(g)}{N_{val}(g) + \epsilon},
\end{equation}
where $\epsilon$ is a small constant introduced to avoid division by zero.If $\rho_g$ exceeds a threshold $\tau_{conf}$, the Gaussian is deemed a geometric artifact, and its label is corrected to background. This mechanism leverages dense multi-view consensus to significantly reduce boundary noise.

\subsection{Multi-View Verification and Spatial Refinement}

While mask-based lifting and background cropping provide initial 3D segmentation, mask inaccuracies introduce artifacts. To resolve this, we enforce multi-view verification and spatial continuity.

\noindent \textbf{Cross-View Back-Projection Verification. } Utilizing the keyframe set $\mathcal{I}_{key}$, we reference the 2D bounding box $\mathcal{B}_{i}$ of the target in each keyframe $I_i \in \mathcal{I}_{key}$ generated by the VLM. If there are multiple instances, we take their spatial union. For each Gaussian $g \in \mathcal{G}_{fg}$, we project it onto keyframe $I_i$ to obtain the 2D pixel coordinates $\mathbf{u}_{g, i}$. We then count the valid observation frequency $N_{vis}(g)$ and the out-of-bounds frequency $N_{out}(g)$ for Gaussian $g$ across the keyframes:
\begin{equation}
	N_{vis}(g) = \sum_{I_i \in \mathcal{I}_{key}} \mathbb{I}_{vis}(g, I_i), \quad N_{out}(g) = \sum_{I_i \in \mathcal{I}_{key}} \mathbb{I}_{out}(g, I_i)
\end{equation}
Here, the visibility indicator $\mathbb{I}_{vis}(g, I_i) = 1$ if $\mathbf{u}_{g, i}$ is within the image viewport. The out-of-bounds indicator $\mathbb{I}_{out}(g, I_i) = 1$ if the point is visible in the viewport but falls outside $\mathcal{B}_{i}$. To avoid over-cropping the target geometry due to single-view occlusions or 2D localization errors, we compute the cross-view out-of-bounds ratio $R_{out}(g) = \frac{N_{out}(g)}{N_{vis}(g) + \epsilon}$ as a robust filtering metric, where $\epsilon$ is a small constant. We introduce a minimum observation threshold $\tau_{views}$ and an out-of-bounds tolerance threshold $\tau_{box}$. If a Gaussian $g$ satisfies $N_{vis}(g) \ge \tau_{views}$ and $R_{out}(g) > \tau_{box}$, it is predominantly outside the bounding box across multiple views. Consequently, we correct it to background; otherwise, it retains its foreground label.

\noindent \textbf{Local Refinement via Spatial Consistency.} Occlusions and mask edge errors often create internal cavities within 3D targets. To reconstruct structural integrity, we apply a 3D K-Nearest Neighbor (KNN) spatial label diffusion algorithm using KD-Trees. For an unlabeled Gaussian $g_{null}$, we query its $k$ nearest neighboring Gaussians. If the fraction of neighbors belonging to the foreground set $\mathcal{G}_{fg}$ exceeds a reliability threshold, i.e.,
\begin{equation}
	\frac{1}{k} \sum_{g_j \in \text{KNN}(g_{null})} \mathbb{I}(g_j \in \mathcal{G}_{fg}) \ge \tau_{knn}
\end{equation}
we treat the Gaussian as a missing internal structure and update its label to foreground. This diffusion mechanism effectively fills local discontinuities, ensuring the final 3D semantic output is structurally coherent and complete.

\section{Experiments}
\subsection{Benchmark}
To advance research on Generalized Referring 3D Gaussian Splatting Segmentation (GR3DGS), we introduce two highly challenging benchmark datasets: \textbf{GR-LERF} and \textbf{GR-ScanNet}. 
For 2D pixel-level alignment, we build \textbf{GR-LERF} on top of LERF scenes~\cite{lerf2023}, and manually curate a large collection of GR3DGS referring instructions to evaluate pixel-level comprehension under the GR3DGS setting. 
However, datasets for assessing point-level understanding in GR3DGS remain scarce. 
To fill this gap, we construct \textbf{GR-ScanNet} based on ScanNet~\cite{dai2017scannet}, adopting the 10 representative indoor scenes used in OpenGaussian~\cite{wu2024opengaussian} and designing complex GR3DGS instructions tailored to these scenes. 
All instructions in both benchmarks are produced via a rigorous manual annotation and cross-validation protocol (see Appendix), ensuring accurate ground-truth supervision.

\subsection{Implementation Details}

We implement ZeroSplat in PyTorch using a single NVIDIA RTX 4090D GPU. For geometry guided keyframe selection, we discretize scenes with a base voxel size of s = 0.1, compute curvature within a radius of $r = \eta s=0.25$ at a resolution of $\gamma s = 0.05$, where $\gamma = 0.5$ and $\eta=2.5$, and rescale surface variation to a saliency interval $[w_{min}, w_{max}]=[1, 10]$ to greedily select $K=30$ keyframes ($\mathcal{I}_{key}$). We deploy qwen3-vl-30b-a3b-instruct on these keyframes to extract labels and generate 2D bounding boxes. We set high-confidence threshold $\tau_{high} = 0.6$ and a base threshold $\tau_{base} = 0.3$, with a quantity threshold of $N_{target} = 30$ and $N_{safe} = 6$. Specifically, we employ a fallback threshold $\tau_{safe} = 0$ for open-vocabulary datasets, while $\tau_{safe} = 0.15$ is used for referring expression datasets. Mask fusion applies a strict threshold $\tau_{merge}=0.8$. Cross view back-Projection Verification necessitates a conflict ratio threshold $\tau_{conf}=0.8$. Cross view background cropping necessitates a minimum observation threshold $\tau_{views}=8$ and an out-of-bounds tolerance threshold $\tau_{box}=0.8$, with $\epsilon$ a small constant set to $10^{-6}$. For geometric refinement, we apply 3D KNN spatial diffusion over $k=40$ neighbors with a reliability threshold $\tau_{knn}=0.8$.

\subsection{Efficiency and Versatility Analysis}
Table~\ref{tab:computation_cost} summarizes the trade-off between task capability and resource consumption across different technical paradigms in 3D semantic understanding, positioning ZeroSplat within the current landscape.
In terms of task capability, most 3D semantic methods are primarily designed for category-level retrieval and are not applicable to language expressions with complex spatial constraints.
While referring segmentation frameworks introduce natural-language interaction, their reasoning is often confined to 2D rendered views and is restricted to the single-target setting, making generalized 3D referring (i.e., handling $0$, $1$, or $N$ instances) challenging.
ZeroSplat fills this gap by enabling generalized referring understanding directly in 3D space.
From an efficiency perspective, prior methods typically rely on scene-specific optimization and incur non-trivial training costs, often accompanied by additional feature storage.
In contrast, ZeroSplat adopts a decoupled, training-free design that requires neither scene optimization nor extra feature storage; specifically, it stores no semantic features per Gaussian and supports plug-and-play inference.
This design offers a practical pathway for deploying 3D semantic understanding on resource-constrained devices.

\begin{table*}[t!]
	\centering
	\small
\caption{Comparison of representative 3D semantic understanding methods by task support and resource cost. ``Referring Task'' indicates language-based referring segmentation (ours supports generalized $0/1/N$). ``Scene Opt.'' denotes per-scene optimization. ``Train Time'', ``Storage'', and ``Peak VRAM'' report per-scene optimization time, extra feature storage beyond 3DGS, and peak GPU memory usage, respectively.}
	\label{tab:computation_cost}
	\renewcommand{\arraystretch}{1.0}
	\setlength{\tabcolsep}{4pt}
	\resizebox{\textwidth}{!}{%
		\begin{tabular}{l c c c c c c c}
			\toprule
			Method & Venue & Domain & Referring Task & Scene Opt. & Train Time & Storage & Peak VRAM \\
			\midrule
			\multicolumn{8}{l}{\textbf{2D pixel-level methods}} \\
			LEGaussians~\cite{shi_language_2024} & CVPR'24 & 2D Pixel & No & Required & $\sim$2h & $\sim$3GB & $\sim$20\,GB \\
			LangSplat~\cite{qin_langsplat_2024} & CVPR'24 & 2D Pixel & No & Required & $\sim$2h & $\sim$3GB & $\sim$20\,GB \\
			Feature-3DGS~\cite{zhou_feature_2024-1} & CVPR'24 & 2D Pixel & No & Required & $\sim$1h & $\sim$3GB & $\sim$26\,GB \\
			GS-Grouping~\cite{ye_gaussian_2024} & ECCV'24 & 2D Pixel & No & Required & $\sim$1h & - & $\sim$28\,GB \\
			GOI~\cite{qu_goi_2024} & MM'24 & 2D Pixel & No & Required & $\sim$1h & - & $\sim$24\,GB \\
			Occam's LGS~\cite{Cheng_2025_BMVC} & BMVC'25 & 2D Pixel & No & None & None & $\sim$3GB & $\sim$12\,GB \\
			3DVLGS~\cite{peng20243d} & ICLR'25 & 2D Pixel & No & Required & $\sim$2h & $\sim$3GB & $\sim$28\,GB \\
			ReferSplat~\cite{ReferSplat} & ICML'25 & 2D Pixel & Yes (single-target) & Required & $\sim$2h & $\sim$3GB & $\sim$28\,GB \\
			
			\midrule
			\multicolumn{8}{l}{\textbf{3D point-level methods}} \\
			OpenGaussian~\cite{wu2024opengaussian} & NeurIPS'24 & 3D Point & No & Required & $\sim$1h & $\sim$3GB & $\sim$22\,GB \\
			InstanceGaussian~\cite{li_instancegaussian_2025} & CVPR'25 & 3D Point & No & Required & $\sim$2h & $\sim$3GB & $\sim$24\,GB \\
			Dr.Splat(Top-40)~\cite{jun-seong_dr_2025} & CVPR'25 & 3D Point & No & None & $\sim$1h & $\sim$3GB & $\sim$24\,GB \\
			LUDVIG~\cite{marrie2025LUDVIG} & ICCV'25 & 3D Point & No & None & None & $\sim$3GB & $\sim$22\,GB \\
			Ours & -- & 3D Point & Yes (generalized) & None & None & 0 & $\sim$10\,GB \\
			\bottomrule
		\end{tabular}%
	}
\end{table*}

\begin{table*}[t!]
	\centering
	\scriptsize
	\caption{Quantitative results on GR-LERF and GR-ScanNet, measured by mIoU.}
	\label{tab:ref_quantitative_r3dgs}
	\renewcommand{\arraystretch}{1.0}
	\setlength{\tabcolsep}{8pt}
	\resizebox{\linewidth}{!}{
		\begin{tabular}{lcccccc}
			\toprule
			& \multicolumn{5}{c}{GR-LERF} & GR-ScanNet \\
			\cmidrule(lr){2-6} \cmidrule(lr){7-7}
			Method & Ramen & Teatime & Figurines & Waldo & Mean & mIoU \\
			\midrule
			\textbf{Pixel-based} & & & & & & \\
			LangSplat~\cite{qin_langsplat_2024}     & 12.7 &  34.6 & 15.6 & 11.4 & 18.6 & - \\
			GOI~\cite{qu_goi_2024}          & 7.5 & \textbf{60.0} & 18.8 & 40.2 & 31.6 & - \\
			LEGaussians~\cite{shi_language_2024} & 4.2 & 2.5 & 2.7 & 4.5 & 3.5 & - \\
			Occam's LGS~\cite{Cheng_2025_BMVC} & 9.8 & 45.7 & 16.3 & 12.9 & 21.2 & - \\
			GS-Grouping~\cite{ye_gaussian_2024}  & 29.1 & 49.2 & 15.6 & 19.3 & 28.3 & - \\
			Feature-3DGS~\cite{zhou_feature_2024-1} & 1.6 & 11.4 & 0.6 & 17.4 & 7.8 & - \\
			ReferSplat~\cite{ReferSplat} & 4.3 & 10.9 & 11.6 & 8.3 & 8.8 & - \\
			\midrule
			\textbf{Point-based} & & & & & & \\
			OpenGaussian~\cite{wu2024opengaussian}        & 7.3 & 10.5 & 4.5 & 9.1 & 7.9 & 19.8 \\
			InstanceGaussian~\cite{li_instancegaussian_2025}    & 6.9 &  6.8 & 11.3 & 14.7 & 9.9 & 24.5 \\
			Dr.Splat(Top-40)~\cite{jun-seong_dr_2025}    &  12.0 & 8.7 &  7.1 & 11.7 &  9.9 & 21.5 \\
			LUDVIG~\cite{marrie2025LUDVIG} & 25.3 & 32.5 & 23.9 & 13.5 & 23.8 & 15.1 \\
			\textbf{Ours} & \textbf{46.5} & 56.3 & \textbf{52.1} & \textbf{48.4} & \textbf{50.8} & \textbf{41.2} \\
			\bottomrule
		\end{tabular}
	}
\end{table*}

\begin{figure*}[t!]
	\centering
	\includegraphics[width=1\linewidth]{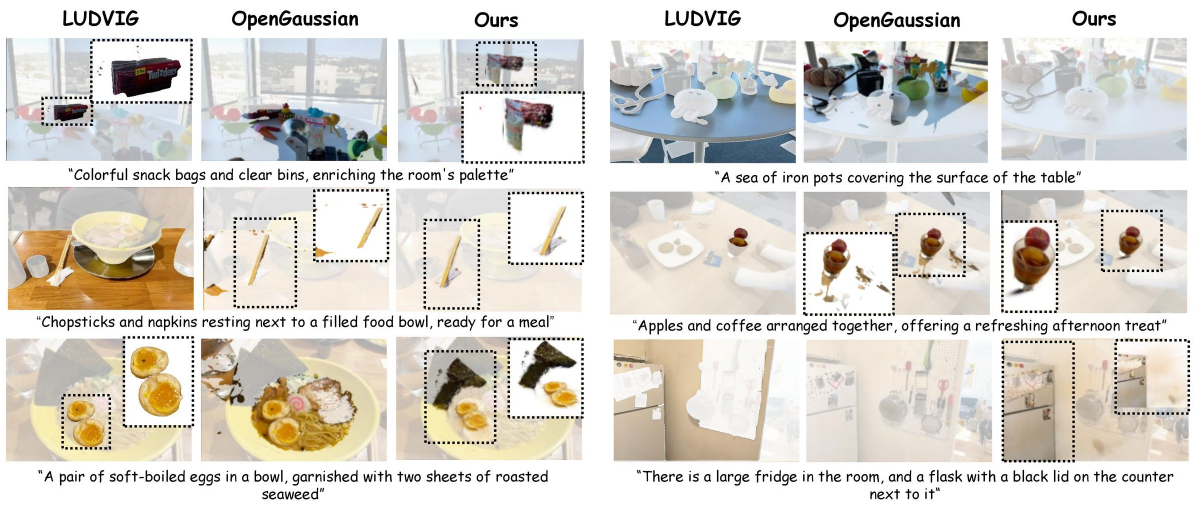}
	\caption{Qualitative results on object selection from the GR-LERF dataset.  }
	\label{003}
\end{figure*}

\begin{table*}[t!]
	\centering
	\scriptsize
	
	\begin{minipage}[t]{0.44\linewidth}
		\centering
		
		\caption{Quantitative results for R3DGS on Ref-LERF measured by mIoU.}
        \vspace{-1em}
		\label{tab:ref_lerf_gr3dgs}
		\renewcommand{\arraystretch}{1.0}
		\setlength{\tabcolsep}{1.3pt}
		\resizebox{\linewidth}{!}{
			\begin{tabular}{l c c c c c}
				\toprule
				Method & Ramen & Teatime & Figurines & Waldo & Mean \\
				\midrule
				Grounded SAM~\cite{ren2024grounded} & 14.1 & 16.9 & 16.0 & 16.2 & 15.8 \\
				LangSplat~\cite{qin_langsplat_2024}    & 12.0 &  7.6 & 17.9 & 17.9 & 13.9 \\
				SPIn-NeRF~\cite{mirzaei2023spin}    &  7.3 & 11.7 &  9.7 & 10.3 &  9.8 \\
				GS-Grouping~\cite{ye_gaussian_2024}  & 27.9 & 14.8 &  8.6 &  6.3 & 14.4 \\
				GOI~\cite{qu_goi_2024}          & 27.1 & 22.9 & 16.5 & 15.7 & 20.5 \\
				ReferSplat~\cite{ReferSplat}    & \textbf{35.2} & \underline{31.3} & \underline{25.7} & \textbf{24.4} & \underline{29.2} \\
				\textbf{Ours} & \underline{30.4} & \textbf{41.4} & \textbf{37.8} & \underline{21.3} & \textbf{32.7} \\
				\bottomrule
			\end{tabular}
		}
		
		\vspace{1.2em} 
		
		\caption{Quantitative results for open-vocabulary 3D semantic segmentation on ScanNet measured by mIoU.} 
        \vspace{-1em}
		\label{tab:scannet_results_revised}
		\renewcommand{\arraystretch}{1.0}
		\setlength{\tabcolsep}{3pt} 
		\resizebox{\linewidth}{!}{
			\begin{tabular}{l ccc}
				\toprule
				Method & 19 cls. & 15 cls. & 10 cls. \\
				\midrule
				OpenGaussian~\cite{wu2024opengaussian} & 24.7 & 30.1 & 38.3 \\
				InstanceGaussian~\cite{li_instancegaussian_2025} & \underline{40.7} & \underline{42.5} & 47.9 \\
				Dr.Splat(Top-40)~\cite{jun-seong_dr_2025} & 29.6 & 38.2 & \textbf{50.8} \\
				LUDVIG~\cite{marrie2025LUDVIG} & 33.9 & 37.4 & 46.4 \\
				\textbf{Ours} & \textbf{44.5} & \textbf{43.7} & \underline{49.7} \\
				\bottomrule
			\end{tabular}
		}
	\end{minipage}\hfill
	\begin{minipage}[t]{0.54\linewidth}
		\centering
		\caption{Quantitative results for open-vocabulary object selection on LERF measured by mIoU.}
		\label{tab1}
		\renewcommand{\arraystretch}{1.2} 
		\setlength{\tabcolsep}{1.3pt} 
		\resizebox{\linewidth}{!}{
			\begin{tabular}{l c c c c c}
				\toprule
				Method & Ramen & Teatime & Figurines & Waldo & Mean \\
				\midrule
				\textbf{Pixel-based} & & & & & \\
				LEGaussians~\cite{shi_language_2024} & 46.0 & 60.3 & 40.8 & 39.4 & 46.6 \\
				LangSplat~\cite{qin_langsplat_2024} & 51.2 & 65.1 & 44.7 & 44.5 & 51.4 \\
				Feature-3DGS~\cite{zhou_feature_2024-1} & 43.7 & 58.8 & 40.5 & 39.6 & 45.7 \\
				GS-Grouping~\cite{ye_gaussian_2024} & 45.5 & 60.9 & 40.0 & 38.7 & 46.3 \\
				GOI~\cite{qu_goi_2024} & 52.6 & 63.7 & 44.5 & 41.4 & 50.6 \\
				ReferSplat & \underline{55.1} & 50.1 & \textbf{67.5} & 48.9 & 55.4 \\
				Occam's LGS~\cite{Cheng_2025_BMVC} & 51.0 & \underline{70.2} & 58.6 & \textbf{65.3} & \underline{61.3} \\
				3DVLGS~\cite{peng20243d} & \textbf{61.4} & \textbf{73.5} & \underline{58.1} & \underline{54.8} & \textbf{62.0} \\
				\midrule
				\textbf{Point-based} & & & & & \\
				OpenGaussian~\cite{wu2024opengaussian} & 31.0 & \underline{60.4} & 39.3 & 22.7 & 38.4 \\
				InstanceGaussian~\cite{li_instancegaussian_2025} & 24.6 & \textbf{63.4} & 45.5 & 29.2 & 40.7 \\
				Dr.Splat(Top-40)~\cite{jun-seong_dr_2025} & 24.7 & 57.2 & 53.4 & 39.1 & 43.6 \\
				LUDVIG~\cite{marrie2025LUDVIG} & \underline{42.3} & 58.6 & \underline{58.0} & \underline{42.8} & \underline{50.4} \\
				\textbf{Ours} & \textbf{42.7} & 57.6 & \textbf{58.1} & \textbf{51.2} & \textbf{52.4} \\
				\bottomrule
			\end{tabular}
		}
	\end{minipage}
\end{table*}

\subsection{Generalized Referring 3D Gaussian Splatting Segmentation}
\noindent\textbf{Settings.}
\textbf{1) Task.} We evaluate GR3DGS, where a model must segment the targets specified by a natural-language instruction. The instruction may correspond to $0$, $1$, or $N$ instances.
\textbf{2) Datasets.} We evaluate on GR-LERF and GR-ScanNet. GR-LERF measures pixel-level segmentation accuracy on 2D rendered views, while GR-ScanNet measures point-level segmentation accuracy in 3D space.
\textbf{3) Baselines.} We compare with representative 3D semantic understanding methods. We use the hyperparameters and model settings reported in the original papers to ensure fair comparisons. Due to representation constraints, some pixel-only methods are not evaluated on GR-ScanNet.

\noindent\textbf{Results on GR-LERF.}
As shown in Table~\ref{tab:ref_quantitative_r3dgs}, ZeroSplat significantly outperforms all baselines in pixel-level segmentation. Qualitatively, as shown in Figure~\ref{003}, prior methods like Opengaussian and LUDVIG struggle with multi-target scenes, often merging or missing nearby instances due to over-smoothed embeddings. In contrast, ZeroSplat uses VLM-based compositional reasoning to accurately segment all text-specified targets, yielding significantly higher recall.

\begin{figure*}[t!]
	\centering
	\includegraphics[width=1\linewidth]{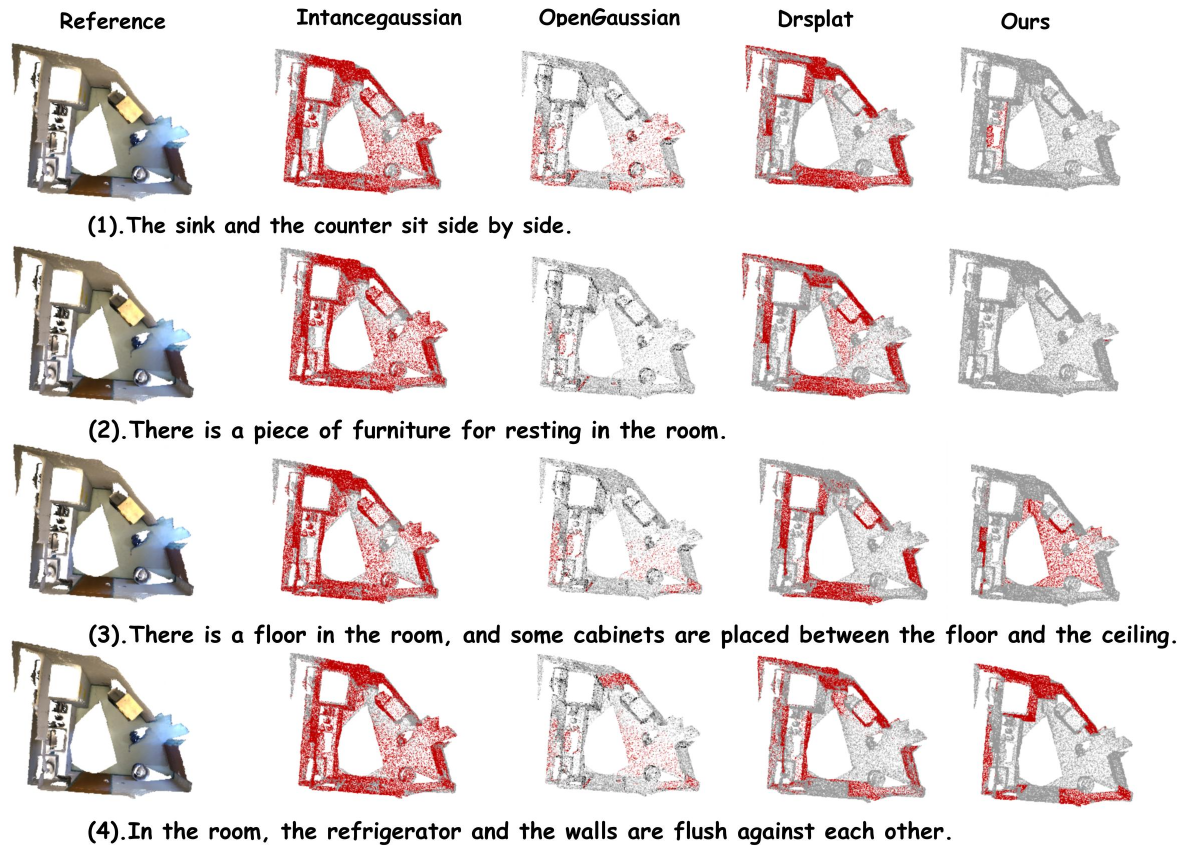}
	\caption{Qualitative results on object selection from the GR-ScanNet dataset. }
	\label{004}
\end{figure*}

\noindent\textbf{Results on GR-ScanNet.}
Quantitatively, ZeroSplat surpasses existing frameworks and establishes a new state-of-the-art for intrinsic 3D point-level segmentation, as shown in Table~\ref{tab:ref_quantitative_r3dgs}. Qualitatively, visual comparisons in Figure~\ref{004} reveal that baseline methods struggle to interpret complex spatial and functional queries. InstanceGaussian and DrSplat exhibit severe over-segmentation, frequently bleeding into irrelevant background regions, whereas OpenGaussian fails to effectively localize targets, resulting in sparse or missing predictions. In contrast, our approach accurately grounds intricate natural language descriptions, isolating the exact queried instances with crisp boundaries and high semantic purity.

\subsection{Referring 3D Gaussian Splatting Segmentation}
\noindent\textbf{Settings.}
\textbf{1) Task.} This task requires localizing and segmenting a single target object in a 3D scene given a complex natural-language instruction. Unlike category-level retrieval, the instruction often specifies spatial relations or fine-grained attributes, which demands strong understanding of 3D structure and object properties.
\textbf{2) Baselines.} We compare ZeroSplat against ReferSplat~\cite{ReferSplat} and several pixel-level understanding methods. Notably, unlike these baselines, our approach distinguishes itself by achieving intrinsic point-level understanding.
\textbf{3) Datasets.} We evaluate on the standard test set released by ReferSplat.

\noindent\textbf{Results.} As shown in Table~\ref{tab:ref_lerf_gr3dgs}, ZeroSplat achieves a strong mean mIoU of 32.7. Notably, although ZeroSplat is entirely training-free, it still outperforms the fully supervised ReferSplat (+3.5 mIoU).

\begin{table*}[t!]
	\centering
	
	\begin{minipage}[t]{0.28\linewidth}
		\centering
		\caption{Ablation of core components on GR-ScanNet.}
		\label{tab:ablation_global}
		\renewcommand{\arraystretch}{0.9}
		\setlength{\tabcolsep}{8pt} 
		\resizebox{\linewidth}{!}{
			\begin{tabular}{ccc}
				\toprule
				VLM & KNN & mIoU \\
				\midrule
				& & 24.7 \\
				\checkmark & & 38.6 \\
				& \checkmark & 26.2 \\
				\checkmark & \checkmark & \textbf{41.2} \\
				\bottomrule
			\end{tabular}
		}
	\end{minipage}
	\hfill 
	\begin{minipage}[t]{0.30\linewidth}
		\centering
		\caption{Ablation of internal VLM mechanisms on GR-ScanNet.}
		\label{tab:ablation_vlm}
		\renewcommand{\arraystretch}{1.3}
		\setlength{\tabcolsep}{3pt}
		\resizebox{\linewidth}{!}{
			\begin{tabular}{l c}
				\toprule
				Method & mIoU \\
				\midrule
				w/o Semantic Label Extraction & 24.3 \\
				w/o Bounding Box Filter & 40.1 \\
				Random Keyframe Sampling & 39.6 \\
				Full VLM Module & \textbf{41.2} \\
				\bottomrule
			\end{tabular}
		}
	\end{minipage}%
	\hfill 
	\begin{minipage}[t]{0.4\linewidth}
		\centering
		\caption{Comparison of different spatial refinement strategies on GR-ScanNet.}
		\label{tab:ablation_refinement}
		\renewcommand{\arraystretch}{1.2} 
		\setlength{\tabcolsep}{10pt}
		\resizebox{\linewidth}{!}{ 
			\begin{tabular}{lc}
				\toprule
				Method & mIoU \\
				\midrule
				w/o Refinement & 38.6 \\
				3D Radius Search & 39.8 \\
				3D KNN Diffusion (Ours) & \textbf{41.2} \\
				\bottomrule
			\end{tabular}
		}
	\end{minipage}
\end{table*}

\subsection{Open-Vocabulary 3D Gaussian Splatting Segmentation}
\noindent\textbf{Settings.}
\textbf{1) Task.} This task localizes and segments a single target object in a 3D scene given a textual category.
\textbf{2) Datasets.} We conduct extensive evaluations on LERF and ScanNet to assess performance from complementary perspectives. Specifically, we use LERF to measure 2D performance by comparing rendered-view segmentations against its 2D annotations, and we use ScanNet to measure 3D performance using its precise point-level semantic labels to directly evaluate 3D localization and Gaussian segmentation.
\textbf{3) Baselines.} We compare against a range of representative open-vocabulary understanding methods. Due to representational limitations, some pixel-only methods are not evaluated on ScanNet. 
\textbf{4) Adaptation.} For this task, we bypass the VLM-driven semantic parsing and geometric anchoring stages, as they are specifically designed for complex referring instructions. Instead, we directly use the category names as semantic labels for our 2D-to-3D lifting and refinement pipeline.

\noindent\textbf{Results.} Although our method is not designed for open-vocabulary semantic segmentation benchmarks, ZeroSplat demonstrates stable and competitive performance. As shown in Tables~\ref{tab:scannet_results_revised} and~\ref{tab1}, ZeroSplat delivers leading results on both datasets: it achieves 52.4 mIoU on LERF, surpassing the previous best by 2.0 mIoU and establishing a new state of the art; on ScanNet, it attains the best performance under both the 19-class and 15-class evaluation protocols. These results suggest that our pipeline generalizes well beyond its original design goal and remains effective even under this simplified category-driven setting.

\subsection{Ablation Study}
To evaluate the contribution of each proposed component, we conduct extensive ablation studies on the GR-ScanNet dataset using the mIoU metric. 

\noindent \textbf{Effectiveness of Key Components.} We validate our core components in Table~\ref{tab:ablation_global}. The baseline mask-lifting strategy yields a sub-optimal 24.7 mIoU due to unconstrained multi-view inconsistencies. Integrating the VLM for spatial anchoring significantly boosts performance to 38.6 mIoU (+13.9), effectively pruning cross-view false positives. Applying the 3D KNN diffusion alone improves structural completeness to 26.2 mIoU (+1.5). Combining both modules yields the best performance (41.2 mIoU), demonstrating that global semantic localization and local geometric refinement are highly complementary.

\noindent \textbf{VLM Parsing and Geometric Anchoring.} Table~\ref{tab:ablation_vlm} ablates the VLM module's mechanisms. Removing semantic label extraction causes a severe performance collapse (to 24.3 mIoU), as feeding free-form prompts into the 2D segmentation model introduces high ambiguity. Replacing geometry-guided keyframe selection with uniform random sampling decreases mIoU to 39.6, highlighting the need for geometric variation to resolve multi-view ambiguities. Discarding the 2D bounding box filter drops performance to 40.1 mIoU, confirming its necessity as a spatial prior for suppressing misclassified background Gaussians.

\noindent \textbf{Spatial Refinement Strategies.} Table~\ref{tab:ablation_refinement} compares strategies for repairing internal cavities caused by view-dependent occlusions. A fixed-radius 3D search struggles (39.8 mIoU) because Gaussian density varies drastically across scenes; a static radius fails to balance sparse and dense regions. In contrast, our 3D KNN diffusion dynamically adapts to local point density, robustly filling structural voids while preserving sharp object boundaries, achieving the peak 41.2 mIoU.

\section{Conclusion}

In this paper, we introduce the Generalized Referring 3D Gaussian Splatting Segmentation (GR3DGS) task. Unlike the conventional single-target paradigm, GR3DGS addresses real-world complexities by handling multiple targets and absent objects. To tackle this task, we propose ZeroSplat, a zero-feature and training-free framework for intrinsic point-level 3D scene understanding. By lifting 2D Vision-Language Model (VLM) priors into 3D space via multi-view geometric constraints and spatial diffusion, ZeroSplat eliminates the need for expensive per-scene optimization and auxiliary semantic feature storage. Furthermore, we construct two benchmarks, GR-LERF and GR-ScanNet, to evaluate performance at both the pixel and point levels. Extensive experiments show that ZeroSplat significantly outperforms existing state-of-the-art methods in generalized, single-target, and open-vocabulary scenarios, while maintaining high efficiency as a plug-and-play solution. We believe our framework and benchmarks will serve as a solid foundation for future research in Embodied AI and interactive 3D scene understanding.

\bibliographystyle{splncs04}
\bibliography{main}

@String(CVPR  = {IEEE Conf. Comput. Vis. Pattern Recog.})

@String(ICCV  = {Int. Conf. Comput. Vis.})

@String(ICML  = {Int. Conf. Mach. Learn.})

@String(BMVC  = {Brit. Mach. Vis. Conf.})

@String(AAAI  = {AAAI})

@String(CVPR  = {CVPR})

@String(ICCV  = {ICCV})

@String(ICML  = {ICML})

@String(BMVC  =	{BMVC})

@article{liang2024supergseg,
  title={Supergseg: Open-vocabulary 3d segmentation with structured super-gaussians},
  author={Liang, Siyun and Wang, Sen and Li, Kunyi and Niemeyer, Michael and Gasperini, Stefano and Navab, Nassir and Tombari, Federico},
  journal={arXiv preprint arXiv:2412.10231},
  year={2024}
}

@inproceedings{wang2022cris,
  title={Cris: Clip-driven referring image segmentation},
  author={Wang, Zhaoqing and Lu, Yu and Li, Qiang and Tao, Xunqiang and Guo, Yandong and Gong, Mingming and Liu, Tongliang},
  booktitle={Proceedings of the IEEE/CVF conference on computer vision and pattern recognition},
  pages={11686--11695},
  year={2022}
}

@inproceedings{he2024refmask3d,
  title={Refmask3d: Language-guided transformer for 3d referring segmentation},
  author={He, Shuting and Ding, Henghui},
  booktitle={Proceedings of the 32nd ACM International Conference on Multimedia},
  pages={8316--8325},
  year={2024}
}

@inproceedings{wu20243d,
  title={3d-stmn: Dependency-driven superpoint-text matching network for end-to-end 3d referring expression segmentation},
  author={Wu, Changli and Ma, Yiwei and Chen, Qi and Wang, Haowei and Luo, Gen and Ji, Jiayi and Sun, Xiaoshuai},
  booktitle={Proceedings of the AAAI Conference on Artificial Intelligence},
  volume={38},
  number={6},
  pages={5940--5948},
  year={2024}
}

@article{liu2024less,
  title={Less: Label-efficient and single-stage referring 3d segmentation},
  author={Liu, Xuexun and Xu, Xiaoxu and Li, Jinlong and Zhang, Qiudan and Wang, Xu and Sebe, Nicu and Ma, Lin},
  journal={Advances in Neural Information Processing Systems},
  volume={37},
  pages={11164--11185},
  year={2024}
}

@inproceedings{liu2017recurrent,
  title={Recurrent multimodal interaction for referring image segmentation},
  author={Liu, Chenxi and Lin, Zhe and Shen, Xiaohui and Yang, Jimei and Lu, Xin and Yuille, Alan},
  booktitle={Proceedings of the IEEE international conference on computer vision},
  pages={1271--1280},
  year={2017}
}

@inproceedings{huang2021text,
  title={Text-guided graph neural networks for referring 3d instance segmentation},
  author={Huang, Pin-Hao and Lee, Han-Hung and Chen, Hwann-Tzong and Liu, Tyng-Luh},
  booktitle={Proceedings of the AAAI conference on artificial intelligence},
  volume={35},
  number={2},
  pages={1610--1618},
  year={2021}
}

@inproceedings{ding2026extrinsplat,
  title={ExtrinSplat: Decoupling Geometry and Semantics for Open-Vocabulary Understanding in 3D Gaussian Splatting},
  author={Ding, Jiayu and Liu, Xinpeng and Pan, Zhiyi and Long, Shiqiang and Li, Ge},
  booktitle={Proceedings of the IEEE/CVF Conference on Computer Vision and Pattern Recognition},
  pages={31019--31028},
  year={2026}
}

@inproceedings{liu2026egotl,
  title={EgoTL: Egocentric Think-Aloud Chains for Long-Horizon Tasks},
  author={Liu, Lulin and Li, Dayou and Liang, Yiqing and Jiang, Sicong and Vijay, Hitesh and Hu, Hezhen and Xu, Xuhai and Liu, Zirui and Shakkottai, Srinivas and Li, Manling and others},
  booktitle={Proceedings of the IEEE/CVF Conference on Computer Vision and Pattern Recognition},
  pages={2017--2027},
  year={2026}
}

@article{li2026egocentric,
  title={Egocentric World Model for Photorealistic Hand-Object Interaction Synthesis},
  author={Li, Dayou and Liu, Lulin and Liu, Bangya and Zhou, Shijie and Feng, Jiu and Lu, Ziqi and Zheng, Minghui and You, Chenyu and Fan, Zhiwen},
  journal={arXiv preprint arXiv:2603.13615},
  year={2026}
}

@inproceedings{chen2026physics,
  title={A Physics-Grounded Benchmark for Multi-Agent Dynamics in World Models},
  author={Chen, Nuo and Liu, Lulin and Li, Zihao and Zeng, Ziyao and Zhu, Zihao and Cong, Wenyan and Hong, Junyuan and Yang, Yunhao and Tu, Zhengzhong and Wang, Yan and others},
  booktitle={The 2nd Workshop on Foundation Models Meet Embodied Agents at CVPR 2026}
}

@article{wu2024rg,
  title={Rg-san: Rule-guided spatial awareness network for end-to-end 3d referring expression segmentation},
  author={Wu, Changli and Ji, Jiayi and Wang, Haowei and Ma, Yiwei and Huang, You and Luo, Gen and Fei, Hao and Sun, Xiaoshuai and Ji, Rongrong and others},
  journal={Advances in Neural Information Processing Systems},
  volume={37},
  pages={110972--110999},
  year={2024}
}

@article{sun2025cags,
  title={Cags: Open-vocabulary 3d scene understanding with context-aware gaussian splatting},
  author={Sun, Wei and Zhou, Yanzhao and Jiao, Jianbin and Li, Yuan},
  journal={arXiv preprint arXiv:2504.11893},
  year={2025}
}

@article{yin2025semantic,
  title={Semantic Consistent Language Gaussian Splatting for Point-Level Open-vocabulary Querying},
  author={Yin, Hairong and Zhan, Huangying and Xu, Yi and Yeh, Raymond A},
  journal={arXiv preprint arXiv:2503.21767},
  year={2025}
}

@article{jiang2025votesplat,
  title={VoteSplat: Hough Voting Gaussian Splatting for 3D Scene Understanding},
  author={Jiang, Minchao and Jia, Shunyu and Gu, Jiaming and Lu, Xiaoyuan and Zhu, Guangming and Dong, Anqi and Zhang, Liang},
  journal={arXiv preprint arXiv:2506.22799},
  year={2025}
}

@article{jin2026himachierarchicalmacromicrolearning,
   title={HiMAC: Hierarchical Macro-Micro Learning for Long-Horizon LLM Agents},
   author={Jin, Hongbo and Zhu, Rongpeng and Ding, Jiayu and Zhang, Wenhao and Li, Ge},
   journal={arXiv preprint arXiv:2603.00977},
   year={2026}
}

@article{jin2026contextguard,
  title={ContextGuard: Structured Self-Auditing for Context Learning in Language Models},
  author={Jin, Hongbo and Wang, Chi and Tang, Haoran and Du, Zhongjing and Jiang, Xu and Tian, Jingqi and Zhang, Qiaoman and Ding, Jiayu},
  journal={arXiv preprint arXiv:2605.26827},
  year={2026}
}

@article{jin2026context,
  title={Context-CoT: Enhancing Context Learning via High-Quality Reasoning Synthesis},
  author={Jin, Hongbo and Zhu, Mingnan and Tian, Jingqi and Jiang, Xu and Du, Zhongjing and Tang, Haoran and Xie, Siyi and Zhang, Qiaoman and Ding, Jiayu},
  journal={arXiv preprint arXiv:2605.25354},
  year={2026}
}

@article{jin2026vistamitigatingsemanticinertia,
   title={VISTA: Mitigating Semantic Inertia in Video-LLMs via Training-Free Dynamic Chain-of-Thought Routing},
   author={Jin, Hongbo and Ding, Jiayu and Xie, Siyi and Luo, Guibo and Li, Ge},
   journal={arXiv preprint arXiv:2505.11830},
   year={2026}
}

@article{
xiao2026promptbased,
title={Prompt-based Adaptation in Large-scale Vision Models: A Survey},
author={Xi Xiao and Yunbei Zhang and Lin Zhao and Yiyang Liu and Xiaoying Liao and Zheda Mai and Xingjian Li and Xiao Wang and Hao Xu and Jihun Hamm and Xue Lin and Min Xu and Qifan Wang and Tianyang Wang and Cheng Han},
journal={Transactions on Machine Learning Research},
issn={2835-8856},
year={2026}
}

@inproceedings{xiao2025visual,
  title={Visual instance-aware prompt tuning},
  author={Xiao, Xi and Zhang, Yunbei and Li, Xingjian and Wang, Tianyang and Wang, Xiao and Wei, Yuxiang and Hamm, Jihun and Xu, Min},
  booktitle={Proceedings of the 33rd ACM International Conference on Multimedia},
  pages={2880--2889},
  year={2025}
}

@article{peng20243d,
  title={3d vision-language gaussian splatting},
  author={Peng, Qucheng and Planche, Benjamin and Gao, Zhongpai and Zheng, Meng and Choudhuri, Anwesa and Chen, Terrence and Chen, Chen and Wu, Ziyan},
  journal={arXiv preprint arXiv:2410.07577},
  year={2024}
}

@inproceedings{Cheng_2025_BMVC,
author    = {Jiahuan Cheng and Jan-Nico Zaech and Luc Van Gool and Danda Pani Paudel},
title     = {Occam’s LGS: An Efficient Approach for Language Gaussian Splatting},
booktitle = {36th British Machine Vision Conference 2025, {BMVC} 2025, Sheffield, UK, November 24-27, 2025},
publisher = {BMVA},
year      = {2025}
}

@inproceedings{marrie2025ludvig,
    title={LUDVIG: Learning-Free Uplifting of 2D Visual Features to Gaussian Splatting Scenes},
    author={Marrie, Juliette and Menegaux, Romain and Arbel, Michael and Larlus, Diane and Mairal, Julien},
    booktitle={Proceedings of the IEEE/CVF International Conference on Computer Vision (ICCV)},
    year={2025}
}

@article{ren2024grounded,
  title={Grounded sam: Assembling open-world models for diverse visual tasks},
  author={Ren, Tianhe and Liu, Shilong and Zeng, Ailing and Lin, Jing and Li, Kunchang and Cao, He and Chen, Jiayu and Huang, Xinyu and Chen, Yukang and Yan, Feng and others},
  journal={arXiv preprint arXiv:2401.14159},
  year={2024}
}

@inproceedings{mirzaei2023spin,
  title={Spin-nerf: Multiview segmentation and perceptual inpainting with neural radiance fields},
  author={Mirzaei, Ashkan and Aumentado-Armstrong, Tristan and Derpanis, Konstantinos G and Kelly, Jonathan and Brubaker, Marcus A and Gilitschenski, Igor and Levinshtein, Alex},
  booktitle={Proceedings of the IEEE/CVF Conference on Computer Vision and Pattern Recognition},
  pages={20669--20679},
  year={2023}
}

@inproceedings{tang2026video,
  title={Video spatial reasoning with object-centric 3d rollout},
  author={Tang, Haoran and Cao, Meng and Liu, Ruyang and Liang, Xiaoxi and Li, Linglong and Li, Ge and Liang, Xiaodan},
  booktitle={Proceedings of the AAAI Conference on Artificial Intelligence},
  volume={40},
  number={11},
  pages={9395--9403},
  year={2026}
}

@article{kim2024openvla,
  title={Openvla: An open-source vision-language-action model},
  author={Kim, Moo Jin and Pertsch, Karl and Karamcheti, Siddharth and Xiao, Ted and Balakrishna, Ashwin and Nair, Suraj and Rafailov, Rafael and Foster, Ethan and Lam, Grace and Sanketi, Pannag and others},
  journal={arXiv preprint arXiv:2406.09246},
  year={2024}
}

@article{li2026lmm,
  title={LMM-Track4D: Eliciting 4D Dynamic Reasoning in LMMs via Trajectory-Grounded Dialogue},
  author={Li, Chaoyue and Xu, Yongxue and Feng, Jie and Ding, Jiayu},
  journal={arXiv preprint arXiv:2605.19390},
  year={2026}
}

@misc{kerbl_3d_2023,
	title     = {{3D} {Gaussian} {Splatting} for {Real}-{Time} {Radiance} {Field} {Rendering}},
	publisher = {arXiv},
	author    = {Kerbl, Bernhard and Kopanas, Georgios and Leimkühler, Thomas and Drettakis, George},
	howpublished = {arXiv preprint arXiv:2308.04079},
	year      = {2023},
	note      = {arXiv:2308.04079},
}

@article{ding20263did,   
    title={3D Instruction Ambiguity Detection},  
    author={Ding, Jiayu and Tang, Haoran and Jin, Hongbo and Gao, Wei and Li, Ge},   
    journal={arXiv preprint arXiv:2601.05991},   
    year={2026} 
}

@article{wu2024opengaussian,
	title   = {Opengaussian: Towards point-level 3d gaussian-based open vocabulary understanding},
	author  = {Wu, Yanmin and Meng, Jiarui and Li, Haijie and Wu, Chenming and Shi, Yahao and Cheng, Xinhua and Zhao, Chen and Feng, Haocheng and Ding, Errui and Wang, Jingdong and others},
	journal = {Advances in Neural Information Processing Systems},
	volume  = {37},
	pages   = {19114--19138},
	year    = {2024}
}

@inproceedings{li_instancegaussian_2025,
	title     = {Instancegaussian: {Appearance}-semantic joint gaussian representation for {3D} instance-level perception},
	shorttitle= {Instancegaussian},
	booktitle = {Proceedings of the {Computer} {Vision} and {Pattern} {Recognition} {Conference}},
	author    = {Li, Haijie and Wu, Yanmin and Meng, Jiarui and Gao, Qiankun and Zhang, Zhiyao and Wang, Ronggang and Zhang, Jian},
	year      = {2025},
	pages     = {14078--14088},
}

@inproceedings{shi_language_2024,
	title     = {Language embedded {3D} gaussians for open-vocabulary scene understanding},
	booktitle = {Proceedings of the {IEEE}/{CVF} {Conference} on {Computer} {Vision} and {Pattern} {Recognition}},
	author    = {Shi, Jin-Chuan and Wang, Miao and Duan, Hao-Bin and Guan, Shao-Hua},
	year      = {2024},
	pages     = {5333--5343},
}

@inproceedings{qin_langsplat_2024,
	title     = {Langsplat: {3D} language gaussian splatting},
	shorttitle= {Langsplat},
	booktitle = {Proceedings of the {IEEE}/{CVF} {Conference} on {Computer} {Vision} and {Pattern} {Recognition}},
	author    = {Qin, Minghan and Li, Wanhua and Zhou, Jiawei and Wang, Haoqian and Pfister, Hanspeter},
	year      = {2024},
	pages     = {20051--20060},
}

@inproceedings{zhou_feature_2024-1,
	title     = {Feature {3Dgs}: {Supercharging} 3d gaussian splatting to enable distilled feature fields},
	shorttitle= {Feature 3dgs},
	booktitle = {Proceedings of the {IEEE}/{CVF} {Conference} on {Computer} {Vision} and {Pattern} {Recognition}},
	author    = {Zhou, Shijie and Chang, Haoran and Jiang, Sicheng and Fan, Zhiwen and Zhu, Zehao and Xu, Dejia and Chari, Pradyumna and You, Suya and Wang, Zhangyang and Kadambi, Achuta},
	year      = {2024},
	pages     = {21676--21685},
}

@inproceedings{ye_gaussian_2024,
	title     = {Gaussian grouping: {Segment} and edit anything in {3D} scenes},
	shorttitle= {Gaussian grouping},
	booktitle = {European conference on computer vision},
	publisher = {Springer},
	author    = {Ye, Mingqiao and Danelljan, Martin and Yu, Fisher and Ke, Lei},
	year      = {2024},
	pages     = {162--179},
}

@inproceedings{qu_goi_2024,
	title     = {{GOI}: {Find} {3D} gaussians of interest with an optimizable open-vocabulary semantic-space hyperplane},
	isbn      = {979-8-4007-0686-8},
	shorttitle= {{GOI}},
	language  = {en},
	booktitle = {Proceedings of the 32nd {ACM} {International} {Conference} on {Multimedia}},
	publisher = {ACM},
	author    = {Qu, Yansong and Dai, Shaohui and Li, Xinyang and Lin, Jianghang and Cao, Liujuan and Zhang, Shengchuan and Ji, Rongrong},
	year      = {2024},
}

@inproceedings{jun-seong_dr_2025,
	title     = {Dr. splat: {Directly} referring {3D} gaussian splatting via direct language embedding registration},
	shorttitle= {Dr. splat},
	booktitle = {Proceedings of the {Computer} {Vision} and {Pattern} {Recognition} {Conference}},
	author    = {Jun-Seong, Kim and Kim, GeonU and Yu-Ji, Kim and Wang, Yu-Chiang Frank and Choe, Jaesung and Oh, Tae-Hyun},
	year      = {2025},
	pages     = {14137--14146},
}

@inproceedings{ReferSplat,
  title={{ReferSplat}: Referring Segmentation in 3D Gaussian Splatting},
  author={He, Shuting and Jie, Guangquan and Wang, Changshuo and Zhou, Yun and Hu, Shuming and Li, Guanbin and Ding, Henghui},
  booktitle={International Conference on Machine Learning (ICML)}
}

@inproceedings{munkberg2022extracting,
	title={Extracting triangular 3d models, materials, and lighting from images},
	author={Munkberg, Jacob and Hasselgren, Jon and Shen, Tianchang and Gao, Jun and Chen, Wenzheng and Evans, Alex and M{\"u}ller, Thomas and Fidler, Sanja},
	booktitle={Proceedings of the IEEE/CVF Conference on Computer Vision and Pattern Recognition},
	pages={8280--8290},
	year={2022}
}

@inproceedings{dai2017scannet,
	title     = {Scannet: Richly-annotated 3d reconstructions of indoor scenes},
	author    = {Dai, Angela and Chang, Angel X and Savva, Manolis and Halber, Maciej and Funkhouser, Thomas and Nie{\ss}ner, Matthias},
	booktitle = {Proceedings of the IEEE conference on computer vision and pattern recognition},
	pages     = {5828--5839},
	year      = {2017}
}

@inproceedings{lerf2023,
	author = {Kerr, Justin* and Kim, Chung Min* and Goldberg, Ken and Kanazawa, Angjoo and Tancik, Matthew},
	title = {LERF: Language Embedded Radiance Fields},
	booktitle = {International Conference on Computer Vision (ICCV)},
	year = {2023},
}
\clearpage 
\vspace{1em} 
\appendix  
\vspace{1em} 
\begin{center}
    \Large\bfseries Supplementary Material
\end{center}

\section{Dataset Construction and Statistics}
Since Generalized Referring 3D Gaussian Splatting (GR3DGS) is a novel task, existing benchmarks are insufficient for a comprehensive evaluation. To facilitate future research and ensure reproducibility, we detail the construction process and statistics of our GR-LERF and GR-ScanNet datasets.

\noindent \textbf{Comparison with Existing Datasets.} 
Compared to existing 3D semantic datasets, our benchmarks advance the field in three key aspects (see Table \ref{tab:dataset_comparison}): \textit{1) Evaluation and task type:} Existing datasets like LERF and ScanNet focus on open-vocabulary segmentation, while referring benchmarks like Ref-LERF are limited to 2D pixel-level evaluation. In contrast, our datasets target generalized 3D referring segmentation. By evaluating directly at the 3D point or Gaussian level, they better reflect a model's spatial and geometric understanding. \textit{2) Language complexity:} Traditional datasets mostly rely on simple category labels or short phrases. Our benchmarks use complex natural language instructions with multiple attribute modifiers and spatial relationships. This requires stronger fine-grained text parsing and cross-modal alignment capabilities. \textit{3) Target quantity generalization:} Current referring datasets typically assume a single target per instruction. We remove this constraint by including zero, single, and multiple targets ($0, 1, N$). This setting is closer to real-world scenarios and tests the model's robustness to multi-target scenes as well as its ability to avoid hallucinations when targets are absent.

\noindent \textbf{Annotation Protocol.} 
We designed a strict three-stage annotation pipeline to ensure data quality. \textit{1) Instruction generation:} Annotators observe multi-view renderings of 3D scenes and write candidate instructions based on four aspects: appearance, spatial relationships, functions, and target counting. \textit{2) Mask annotation:} For each instruction, annotators manually create the corresponding 3D target mask. \textit{3) Quality verification:} A three-person expert panel reviews all annotations. Only samples that pass this consistency check are included in the final dataset.

\noindent \textbf{Dataset Statistics.} 
GR-LERF contains 4 3D scenes with 111 instructions, and GR-ScanNet contains 10 3D scenes with 83 instructions. In GR-LERF, the instructions for zero targets (testing refusal and anti-hallucination), single targets (fine-grained localization), and multiple targets (counting and set understanding) are 25, 39, and 47, accounting for 22.52\%, 35.14\%, and 42.34\%, respectively. Similarly, the counts for these three categories ($0, 1, N$) in GR-ScanNet are 9, 21, and 53. This distribution requires models to learn true 3D vision-language alignment rather than relying on dataset bias.

\begin{table*}[t]
\centering
\caption{\textbf{Comparison of 3D semantic benchmarks.} Unlike traditional benchmarks that use simple category tags or restrict queries to single targets ($N=1$), our datasets introduce complex natural language instructions at the 3D point level and support zero, single, and multiple targets ($0, 1, N$). Open-Vocab Seg.: Open-Vocabulary Segmentation; Ref. Seg.: Referring Segmentation; Gen.: Generalized; Rep.: Representation; Inst.: Instructions.}
\label{tab:dataset_comparison}
\resizebox{\linewidth}{!}{
\begin{tabular}{llccccc}
\toprule
\textbf{Dataset} & \textbf{Source} & \textbf{Task Type} & \textbf{Domain} & \textbf{Target Setup} & \textbf{Scenes} & \textbf{Query Scale} \\
\midrule
LERF & LangSplat & Open-Vocab Seg. & 2D & Single Category & 4 & 208 Words \\
ScanNet & OpenGaussian & Open-Vocab Seg. & 3D & Single Category & 10 & 84 Words \\
Ref-LERF & ReferSplat & 2D Ref. Seg. & 2D & Single Target  & 4 & 63 Inst. \\
\rowcolor{gray!15} \textbf{GR-LERF (Ours)} & \textbf{ZeroSplat} & \textbf{Gen. Ref. Seg.} & \textbf{2D} & \textbf{Generalized} & \textbf{4} & \textbf{111 Inst.} \\
\rowcolor{gray!15} \textbf{GR-ScanNet (Ours)} & \textbf{ZeroSplat} & \textbf{Gen. Ref. Seg.} & \textbf{3D} & \textbf{Generalized } & \textbf{10} & \textbf{83 Inst.} \\
\bottomrule
\end{tabular}
}
\end{table*}

\begin{table}[t!]
  \centering
  \footnotesize
\setlength{\tabcolsep}{10pt} 
  \renewcommand{\arraystretch}{0.85}
  \caption{Time cost breakdown for a single query on the GR-LERF Teatime scene. The table details the average processing time per instruction.}
  \label{tab:time_cost}
  \begin{tabular}{lc}
    \toprule
    \textbf{Pipeline Stage} & \textbf{Average Time (s)} \\
    \midrule
    Geometric Keyframe Selection & 224 \\
    VLM Semantic Parsing & 407 \\
    2D Mask Extraction & 480 \\
    2D-to-3D Lifting \& Filtering \& Spatial Refinement & 1195 \\
    \textbf{Total} & \textbf{2306} \\
    \bottomrule
  \end{tabular}
\end{table}

\begin{table}[htbp!]
\centering
\scriptsize
\setlength{\tabcolsep}{3pt}
\renewcommand{\arraystretch}{0.85}
\caption{Time cost breakdown for a single query on the Teatime scene from GR-LERF dataset.}
\label{tab:time_single_cost}
\begin{tabular}{lccccc}
\toprule
\textbf{Method} & \textbf{Data Prep (s)} & \textbf{Train (s)} & \textbf{Lift (s)} & \textbf{Inf / Query (s)} & \textbf{Total (s)} \\
\midrule
OpenGaussian & 5243 & 3993 & 0 & 0.02 & 9237 \\
InstanceGaussian & 5243 & 8518 & 0 & 0.04 & 10333 \\
ReferSplat & 1813 & 5117 & 0 & 0.18 & 6943 \\
DrSplat & 5243 & 4572 & 3186 & 0.08 & 13004 \\
LUDIG & 947 & 0 & 32 & 2.94 & 1085 \\
\textbf{Ours} & 1111 & 0 & 33 & 0 & 1144 \\
\bottomrule
\end{tabular}
\end{table}

\section{Time Efficiency Analysis}
To demonstrate that ZeroSplat is an efficient plug-and-play solution, we provide a detailed time breakdown of its inference pipeline. All evaluations are conducted on a single NVIDIA RTX 4090D GPU. Unlike existing methods that require hours of per-scene optimization, our approach is entirely training-free. Table \ref{tab:time_cost} details the average time spent on each stage. Specifically, our complete pipeline processes the Teatime scene in approximately 38 minutes. 
This runtime demonstrates a significant advantage over mainstream methods. For instance, on the same Teatime scene, InstanceGaussian requires approximately 140 minutes for the 3D Gaussian training phase alone, excluding any subsequent semantic parsing overhead. In addition, We report the minimum runtime from receiving a query to producing the segmen tation result on the Teatime scene under identical settings, as listed in Table \ref{tab:time_single_cost}. Preprocessing denotes the time for extracting 2D scene masks and features, training denotes model optimization, and lifting denotes mapping 2D masks to 3D representations. Inference measures query-field similarity computation, which our method does not require, giving an inference time of 0.

\begin{table*}[t!]
    \centering
    
    \begin{minipage}{\linewidth}
        \centering
        \caption{Impact of different VLM representations on GR-LERF. }
        \label{tab:ablation_vlm_rep}

        \renewcommand{\arraystretch}{1.2} 

        \setlength{\tabcolsep}{14pt} 

        \resizebox{0.92\linewidth}{!}{
        \begin{tabular}{lcc}
            \toprule
            \textbf{VLM Backbone} & \textbf{Open-Source} & \textbf{mIoU} \\
            \midrule
            Gemini-3.1-Flash-Lite & No & 54.3\\
            Qwen3-VL-Plus & No & 56.7 \\
            Qwen3-VL-8B-Instruct & Yes & 49.4 \\
            Qwen3-VL-30B-A3B-Instruct (Ours) & Yes & 50.8 \\
            \bottomrule
        \end{tabular}
        }
    \end{minipage}
    \vspace{1.5em}

    \centering
    \caption{Hyperparameter sensitivity results on the waldo-kitchen scene from the GR-LERF dataset.}
    \label{tab:sensitivity}
    \renewcommand{\arraystretch}{1.2}
    \setlength{\tabcolsep}{6pt}
    \begin{tabular}{l|l}
    \hline
    Parameter (Default) & Parameter Variations $\rightarrow$ mIoU \\
    \hline
    $\tau_{\text{high}}$ (0.6) & 0.8 (48.8), \textbf{0.6 (48.4)}, 0.4 (48.5), 0.2 (48.0), 0.1 (46.1) \\
    $\tau_{\text{base}}$ (0.3) & 0.6 (47.9), 0.4 (49.1), \textbf{0.3 (48.4)}, 0.2 (48.2), 0.1 (43.4) \\
    $\tau_{\text{safe}}$ (0.15) & 0.4 (48.4), \textbf{0.15 (48.4)}, 0.05 (43.6), 0.02 (29.3), 0 (29.3) \\
    $N_{\text{target}}$ (30) & 90 (48.1), 50 (48.3), \textbf{30 (48.4)}, 20 (48.3), 8 (47.7) \\
    $N_{\text{safe}}$ (6) & 20 (48.4), 15 (48.4), 10 (48.4), \textbf{6 (48.4)}, 4 (48.4) \\
    $\tau_{\text{merge}}$ (0.8) & 0.9 (47.8), \textbf{0.8 (48.4)}, 0.6 (48.4), 0.4 (48.8), 0.3 (47.5) \\
    $\tau_{\text{box}}$ (0.8) & 0.9 (48.4), \textbf{0.8 (48.4)}, 0.7 (48.4), 0.6 (48.3), 0.5 (48.1) \\
    $\tau_{\text{views}}$ (8) & 24 (48.3), 15 (48.3), \textbf{8 (48.4)}, 4 (47.7) \\
    $k$ (40) & 80 (48.5), 60 (48.6), \textbf{40 (48.4)}, 20 (48.5), 10 (48.4) \\
    $\tau_{\text{knn}}$ (0.8) & 0.9 (48.7), \textbf{0.8 (48.4)}, 0.7 (48.4), 0.5 (48.5), 0.3 (47.9) \\
    $\tau_{\text{conf}}$ (0.8) & 0.9 (48.2), \textbf{0.8 (48.4)}, 0.6 (51.0), 0.4 (51.9), 0.2 (48.1) \\
\bottomrule
\end{tabular}
\label{tab:hyper_sensitivity_waldo}
\end{table*}

\section{Additional Ablation Studies}

\textbf{Impact of VLM Representation.} 
The core of our feature extraction module is a Vision-Language Model (VLM) . To analyze how VLM representation affects overall performance, we evaluate our method using three different pre-trained VLMs on the GR-LERF dataset. As shown in Table \ref{tab:ablation_vlm_rep}, segmentation accuracy positively correlates with the VLM's representation capability. Models with stronger vision-language alignment consistently yield higher mIoU scores. This demonstrates that the quality of semantic features provided by the VLM dictates task performance. Consequently, the performance ceiling of our approach is not static; it will naturally increase as foundation VLMs evolve.

\begin{figure*}[t]
\centering
\includegraphics[width=\linewidth]{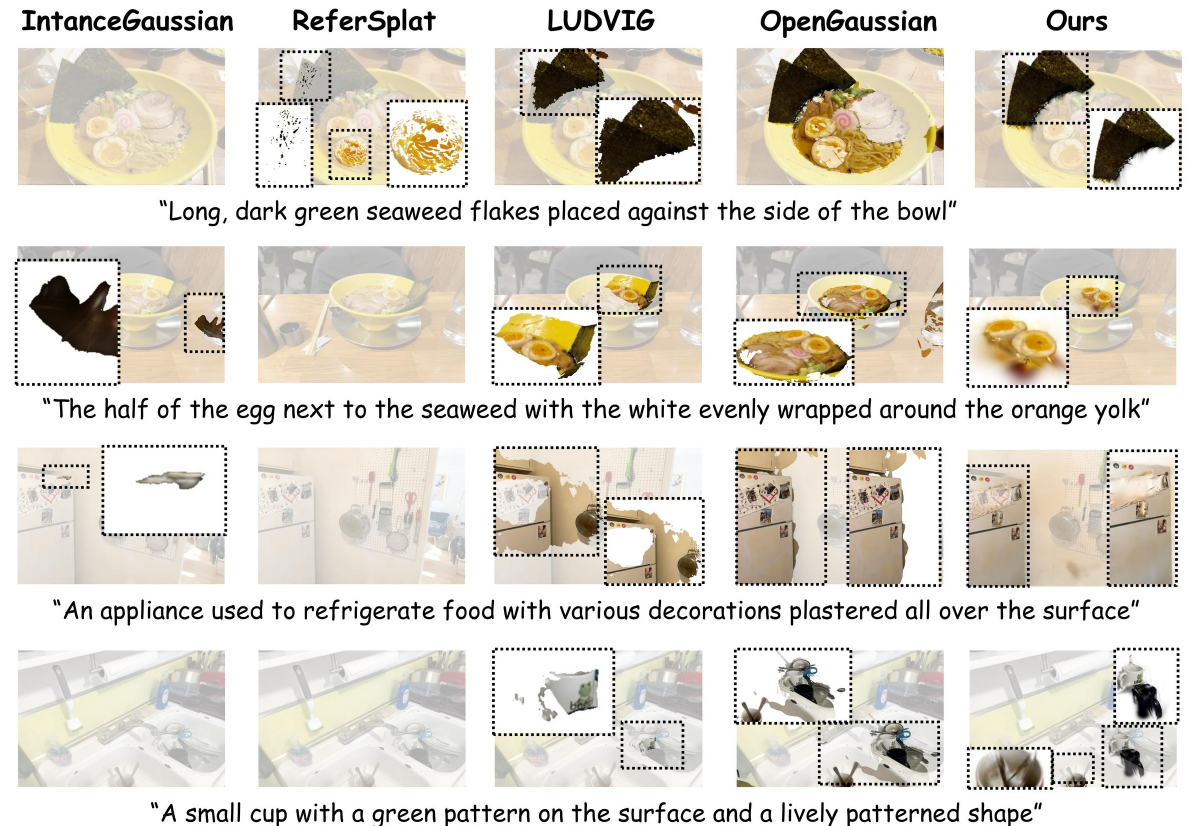}
\caption{Qualitative comparisons on the Ref-LERF dataset.}
\label{fig:qual_reflerf}
\end{figure*}

\begin{figure*}[t]
\centering
\includegraphics[width=\linewidth]{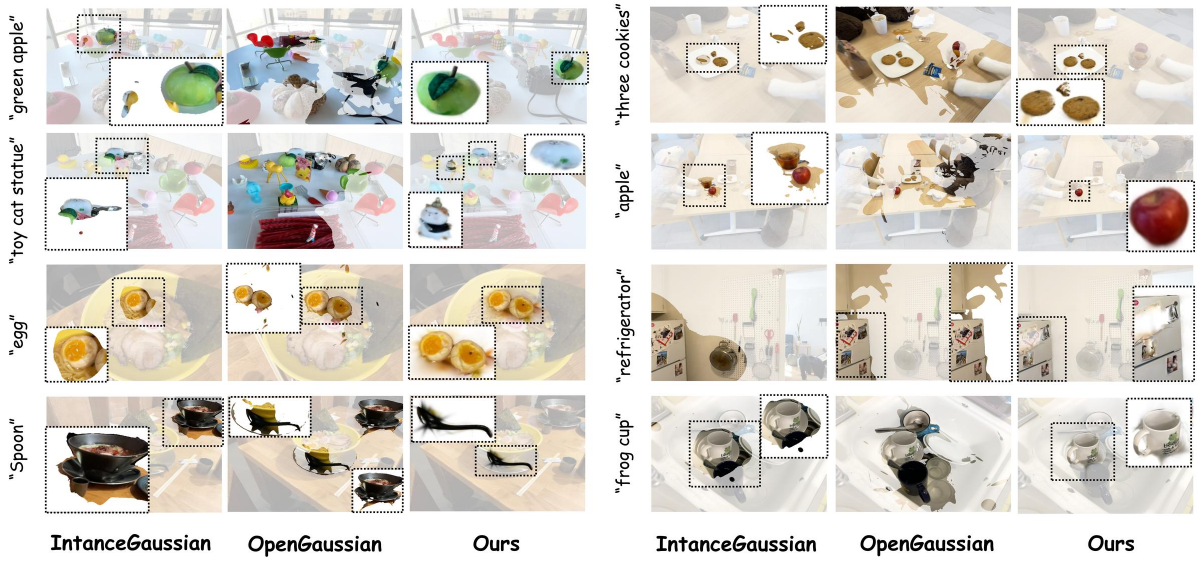}
\caption{Qualitative comparisons on the LERF dataset.}
\label{fig:qual_lerf}
\end{figure*}

\begin{figure*}[t]
\centering
\includegraphics[width=\linewidth]{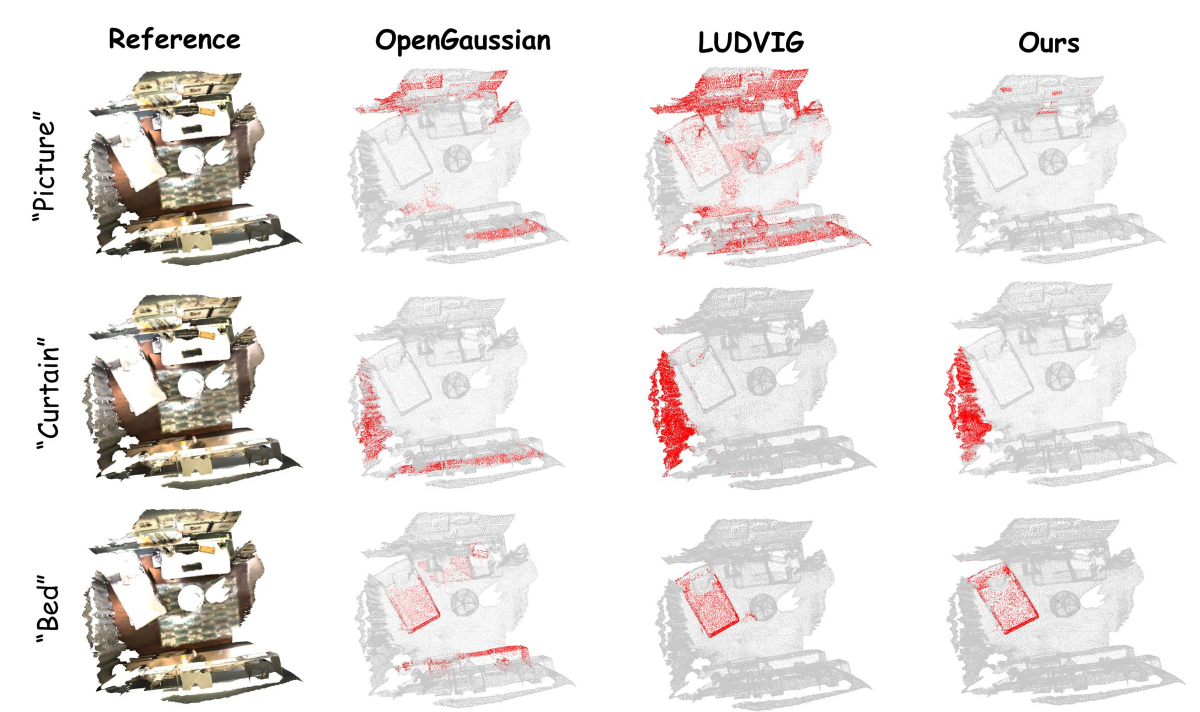}
\caption{Qualitative comparisons on the ScanNet dataset.}
\label{fig:qual_scannet}
\end{figure*}

\noindent \textbf{Hyperparameter Sensitivity.} 
Our framework has several parameters, and they have clear geometric or statistical meanings, such as view counts. The model achieves consistent performance improvements across various datasets without scene-wise parameter tuning. As shown in Table \ref{tab:sensitivity}, we conduct hyperparameter sensitivity analysis on the waldo-kitchen scene from the GR-LERF dataset to further verify robustness. The mIoU remains stable over wide ranges, with degradation only under logical extremes, such as $\tau_{\text{safe}} \to 0$, which disables the safety mechanism.

\section{More Qualitative Results}

To further demonstrate the superiority of our approach, we provide comprehensive qualitative comparisons against state-of-the-art baselines on the LERF, ScanNet, and Ref-LERF datasets. 

\noindent \textbf{Qualitative analysis on Ref-LERF} Figure \ref{fig:qual_reflerf} highlights performance on complex referring expressions involving long descriptive sentences and spatial relationships. Baselines generally fail to comprehend deep semantic contexts, leading to fragmented masks or incorrect localizations. Our method exhibits robust cross-modal alignment. Driven by our Hierarchical VLM Reasoning stage, we first extract concise semantic labels from verbose descriptions and then ground them into precise spatial priors. This two-stage geometry-guided parsing ensures that complex natural language instructions are accurately mapped to the exact 3D visual instances without ambiguity.

\noindent \textbf{Qualitative analysis on LERF} Figure \ref{fig:qual_lerf} illustrates open-vocabulary segmentation results involving fine-grained attributes and multiple targets. Baselines such as InstanceGaussian and OpenGaussian frequently suffer from over-segmentation. They incorrectly include background context, such as the table surface beneath the "green apple" or "three cookies". By contrast, our method accurately localizes targets and consistently produces tight boundaries. This improvement directly benefits from our Cross-View Background Cropping module. By enforcing spatial constraints from VLM-generated 2D bounding boxes across keyframes, we effectively prune out-of-bounds geometric artifacts and prevent background inclusion.

\noindent \textbf{Qualitative analysis on ScanNet} Figure \ref{fig:qual_scannet} presents evaluations on complex 3D indoor scenes. When queried with structural objects like "picture", "curtain", and "bed", existing methods such as OpenGaussian and LUDVIG exhibit severe spatial noise. Their semantic predictions bleed heavily into adjacent walls and floors. Our method significantly mitigates this geometric ambiguity. As detailed in our Cross-View Back-Projection Verification and Local Refinement modules, we leverage dense multi-view consensus to filter conflict points and apply KNN spatial diffusion to fill internal cavities. This mechanism generates clean 3D boundaries that strictly adhere to the physical geometry of the scene.
\begin{figure}[t]
    \centering

    \includegraphics[width=\linewidth]{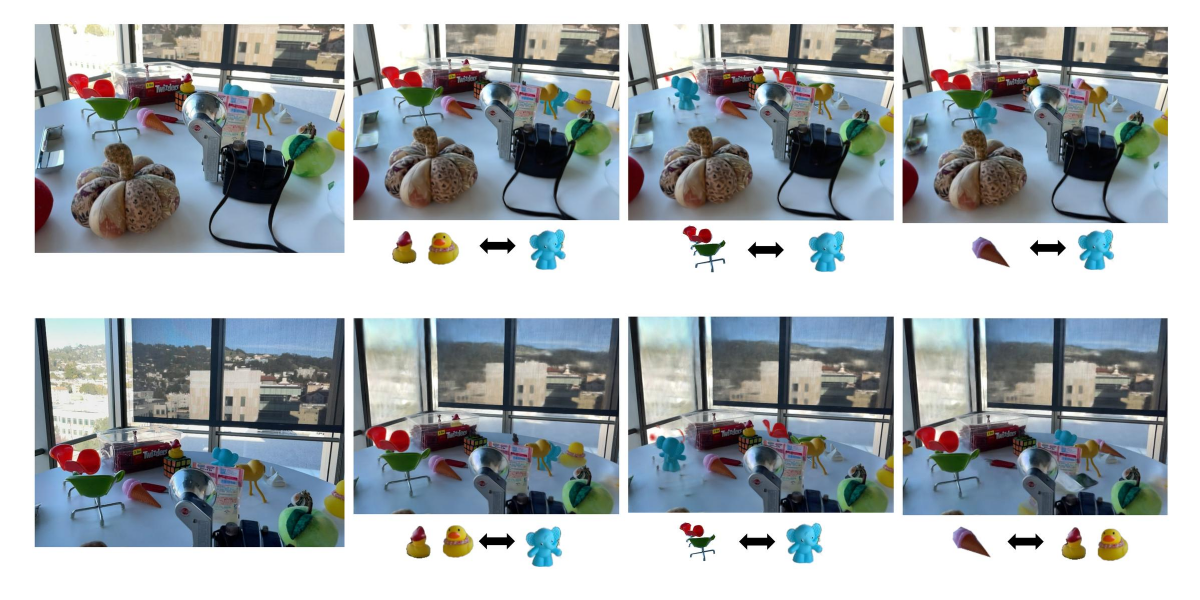}
    \caption{Spatial translation and swapping. Modifying the center coordinates of target Gaussians enables seamless translation and swapping of independent objects.}
    \label{fig:app_translation}

    \vspace{1em}

    \includegraphics[width=\linewidth]{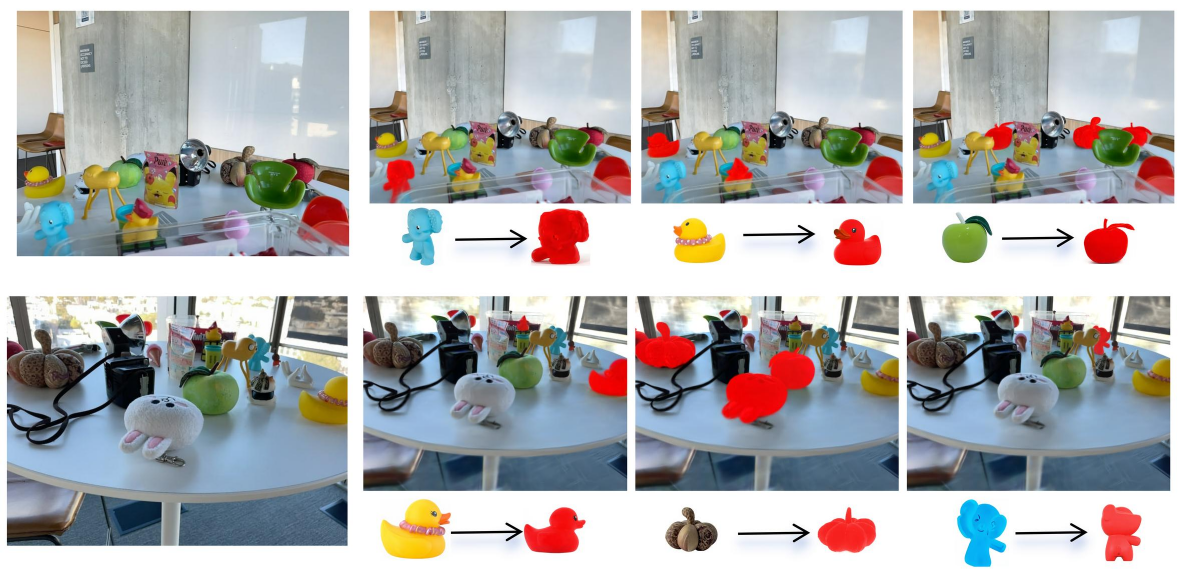}
    \caption{Instance-level appearance stylization. Adjusting the spherical harmonics (SH) coefficients of selected Gaussians enables high-fidelity object color editing.}
    \label{fig:app_stylization}

    \vspace{1em} 

    \includegraphics[width=\linewidth]{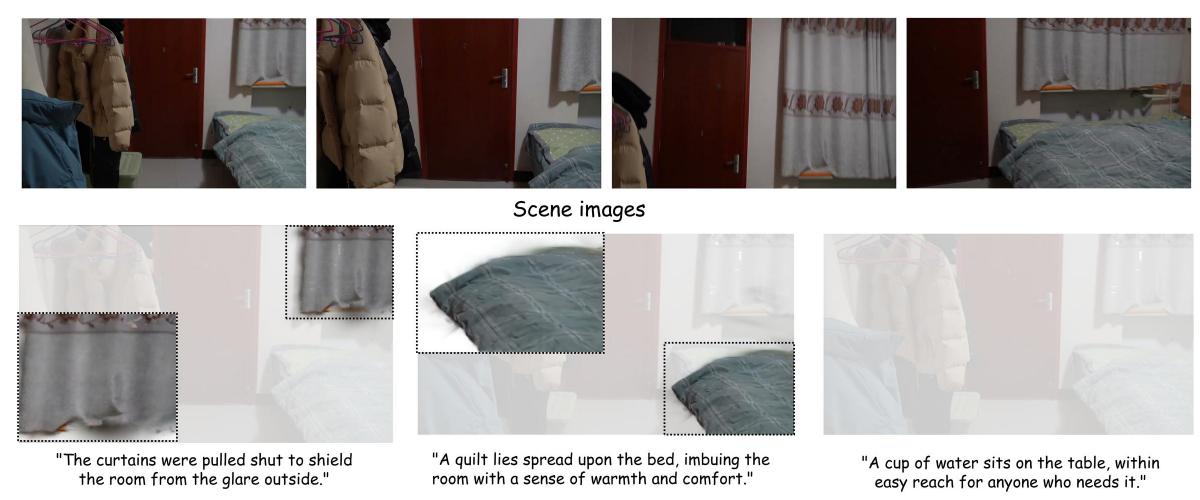}
    \caption{Qualitative results on real-world smartphone captures. ZeroSplat robustly handles complex lighting, occlusions, and long-form descriptive instructions.}
    \label{fig:real_world}
\end{figure}

\section{Failure Analysis}

For challenging materials, such as transparent objects under strong lighting or reflective metallic surfaces in the figurines scene, single-frame SAM3 or VLM errors are often corrected by Adaptive View Selection (Eq.4), since most alternate  views remain reliable. Failures mainly occur under severe VLM limitations:(1) dense clutter disrupts 2D localization; (2) wrong semantic labels from the referring text mislead the pipeline.

\section{Downstream Applications}

ZeroSplat supports highly flexible open-vocabulary 3D scene editing. Traditional 2D image inpainting methods often suffer from multi-view inconsistencies. In contrast, our approach directly decouples and manipulates the attributes of 3D Gaussian primitives in 3D space using natural language instructions.

\noindent \textbf{Spatial translation and swapping.} Because 3D Gaussian primitives possess explicit spatial coordinates, our method enables physical-level spatial transformations of specific instances. As illustrated in Figure \ref{fig:app_translation}, after accurately segmenting the target object, we can translate it to a new location or swap the spatial positions of two independent objects simply by modifying their center coordinates. This editing process causes no geometric interference to unselected objects in the scene.

\noindent \textbf{Instance-level appearance stylization.} After accurately locating the Gaussian primitives of the target instance, we can independently modify its appearance attributes without altering its geometric structure. As shown in Figure \ref{fig:app_stylization}, by directly adjusting the spherical harmonics (SH) color coefficients of the selected Gaussians, we can change the color or style of specific objects.

\section{Real-World Scene Generalization}

To evaluate the generalization capability of ZeroSplat in unconstrained environments, we captured a complex indoor scene using a smartphone and reconstructed its 3D Gaussian representation. As illustrated in Figure \ref{fig:real_world}, our method exhibits remarkable robustness in this unseen environment, which is characterized by complex lighting and severe occlusions. Furthermore, the model accurately parses long-form, context-rich instructions. Notably, when presented with queries for objects absent from the scene (e.g., "A cup of water sits on the table..."), ZeroSplat correctly yields an empty prediction. This demonstrates its strong resistance to object hallucination and reliable zero-target generalization in real-world applications.

\section{Implementation Details and Dataset Configurations}

\subsection{Zero-target evaluation protocol}
To strictly evaluate the model's anti-hallucination capability, we enforce a binary scoring rule for samples where the ground truth contains no target objects (i.e., zero targets). If the model correctly outputs an all-zero mask, the Intersection over Union (IoU) is assigned a value of 1.0. Conversely, if the model predicts any non-zero mask (indicating a false positive localization), the IoU is strictly penalized and assigned 0.0.

\subsection{Geometry-Guided VLM Prompt Design} 

During the hierarchical VLM reasoning stage, we employ a two-stage prompt mechanism to extract semantic labels and 2D bounding boxes. To accommodate the generalized ($0, 1, N$) target settings of our task, we establish strict rules for spatial reasoning and target absence handling within the prompts.

\noindent \textbf{Stage 1: Semantic label extraction.} The first stage converts free-form referring expressions into specific object categories. We require the VLM to perform global spatial reasoning and abstract the targets into a standardized "Color + Noun" format. The exact prompt template is detailed in Figure \ref{fig:prompt_stage1}.

\noindent \textbf{Stage 2: 2D geometric localization.} The second stage utilizes the extracted labels and the original text to localize 2D bounding boxes. To strictly handle the zero-target scenario, we explicitly instruct the model to return -1 when the target is completely absent. The detailed prompt is provided in Figure \ref{fig:prompt_stage2}.

\begin{figure}[t]
\centering
\begin{tcolorbox}[colback=gray!10, colframe=gray!40, arc=2mm, boxrule=0.5pt, left=2mm, right=2mm, top=2mm, bottom=2mm]
\small\ttfamily
System Instruction:\\
You are an expert in visual semantic conversion and scene understanding. I will provide images and a list of 'Referring Expressions' (long descriptions). Your task is to identify ALL object categories mentioned in each expression. An expression may refer to a single object or multiple distinct objects; you must make a comprehensive judgment independently based on the context and ALL provided images.

STRICT RULES:\\
1. Output valid JSON: \{"original expression": ["category1", "category2", ...]\}.\\
2. SPATIAL \& CONTEXTUAL TARGETING: Use spatial relationships (e.g., 'in front of', 'next to', 'on the top') to locate the specific target object(s) in the image. You must analyze the image(s) to ensure the identified object is the one intended by the text.\\
3. ABSTRACTION (Color + Noun): Convert the identified target object into a simplified 'Color + Noun' phrase based on its actual appearance in the images.\\
   Examples:\\
   - Input: 'A round object with a smooth surface directly in front of the white doll' -> If the object in the image is a blue cup, Output: ['blue cup']\\
   - Input: 'fire alarm and smoke detector' -> Output: ['red fire alarm', 'white smoke detector']\\
   - Input: 'the seating furniture on the left' -> If it's a brown chair, Output: ['brown chair']\\
4. QUANTITY: If one expression refers to multiple objects, return them all in the array.\\
5. Do NOT output markdown formatting (```json), just the raw JSON string.\\
6. COMPLETENESS: Return a mapping for EVERY expression in the input list.

User Prompt:\\
Here is the list of descriptions to convert:\\
\{input\_str\}

Please map each description to its specific object category noun:
\end{tcolorbox}
\caption{Stage 1 Prompt: Semantic label extraction. The prompt enforces spatial reasoning and standardizes natural language descriptions into a "Color + Noun" format.}
\label{fig:prompt_stage1}
\end{figure}

\begin{figure}[t]
\centering
\begin{tcolorbox}[colback=gray!10, colframe=gray!40, arc=2mm, boxrule=0.5pt, left=2mm, right=2mm, top=2mm, bottom=2mm]
\small\ttfamily
System Instruction:\\
You are an expert in image localization and visual grounding. I will provide you with MULTIPLE images and a single referring text. Your task is to precisely locate ALL target objects described by the text within the images.

CORE REASONING INSTRUCTIONS:\\
1. VISUAL \& SPATIAL CUES: Strictly rely on spatial relations, color descriptors, and object attributes. If the text describes 'A round object with a smooth surface directly in front of the white doll', you must first identify the white doll as an anchor and then locate the specific object(s) in front of it based on spatial context.\\
2. CROSS-IMAGE SPATIAL REASONING: Analyze ALL provided images to maintain consistency. Use global spatial reasoning to determine if an object in the current frame is indeed part of the target group described by the text.\\
3. ABSENCE OF TARGET: Only return -1 if the target object(s) are absolutely not present.

IMPORTANT RULES:\\
1. Coordinates must be normalized [0, 1000], order: [xmin, ymin, xmax, ymax]. Use the key "box" for the coordinates.

User Prompt:\\
Detect ALL instances of '\{target\_text\}' (Categories: \{short\_names\}) in this image.\\
This text may refer to MULTIPLE object types. Detect ALL of them.\\
Return format (use ONLY these exact keys, output raw JSON only):\\
- MULTIPLE objects: [\{"box": [xmin, ymin, xmax, ymax], "score": 1.0, "label": "object1"\}, ...]\\
- SINGLE object: \{"box": [xmin, ymin, xmax, ymax], "score": 1.0, "label": "object"\}\\
RULES:\\
1. Each object MUST use key "box" (NOT bbox, NOT bbox\_2d) for coordinates.\\
2. Coordinates normalized [0, 1000], order: [xmin, ymin, xmax, ymax]\\
3. Use structural boundaries to distinguish objects\\
4. Return -1 ONLY if absolutely not present\\
5. 'label' field must describe each detected instance
\end{tcolorbox}
\caption{Stage 2 Prompt: 2D geometric localization. This prompt guides the VLM to perform cross-image spatial reasoning and strictly handle the zero-target scenario by outputting -1.}
\label{fig:prompt_stage2}
\end{figure}

\subsection{SAM3 Implementation Details}
To extract 2D semantic masks, we employ the Segment Anything Model 3 (SAM3). SAM3 natively supports promptable segmentation for images and videos. In our pipeline, we directly input the VLM-derived semantic labels into SAM3's text-to-mask interface to generate zero-shot binary masks for the queried targets.

\subsection{ScanNet configuration for open-vocabulary segmentation} 
To ensure a fair comparison on the standard open-vocabulary 3D segmentation task, our evaluation setup strictly aligns with OpenGaussian~\cite{wu2024opengaussian}. We evaluate our model on the exact same 10 scenes selected from the ScanNet dataset. Furthermore, we adopt their identical categorization protocol, structuring the queries into three specific sets. The 19-category set includes wall, floor, cabinet, bed, chair, sofa, table, door, window, bookshelf, picture, counter, desk, curtain, refrigerator, shower curtain, toilet, sink, and bathtub. The 15-category set excludes picture, refrigerator, shower curtain, and bathtub. The 10-category set further excludes cabinet, counter, desk, curtain, and sink.

\end{document}